\documentclass[runningheads]{llncs}
\usepackage{graphicx}
\usepackage{comment}
\usepackage{amsmath,amssymb} 
\usepackage{color}
\usepackage{subfigure}
\usepackage{caption}
\usepackage{hyperref}

\newcommand*\circled[1]{\raisebox{.5pt}{\textcircled{\raisebox{-1.1pt} {#1}}}}

\usepackage{epstopdf}

\usepackage{booktabs}
\usepackage{bbold}

\usepackage[inline]{enumitem}

\usepackage{algorithmic}
\usepackage[linesnumbered,ruled,vlined]{algorithm2e}
\renewcommand{\vec}{\mathbf}

\SetKwInput{KwInput}{Input}                
\SetKwInput{KwOutput}{Output}              

\begin{document}
\pagestyle{headings}
\mainmatter
\def\ECCVSubNumber{188}  

\title{The Group Loss for Deep Metric Learning} 

\titlerunning{The Group Loss for Deep Metric Learning}
%
\author{Ismail Elezi\inst{1} \and
Sebastiano Vascon\inst{1} \and
Alessandro Torcinovich\inst{1} \and
Marcello Pelillo\inst{1} \and
Laura Leal-Taix\'{e}\inst{2}}
\authorrunning{I. Elezi et al.}
%
\institute{Ca' Foscari University of Venice \and Technical University of Munich}
\maketitle

\begin{abstract}
Deep metric learning has yielded impressive results in tasks such as clustering and image retrieval by leveraging neural networks to obtain highly discriminative feature embeddings, which can be used to group samples into different classes.
Much research has been devoted to the design of smart loss functions or data mining strategies for training such networks. 
Most methods consider only pairs or triplets of samples within a mini-batch to compute the loss function, which is commonly based on the distance between embeddings. 
We propose {\it Group Loss}, a loss function based on a differentiable label-propagation method that enforces embedding similarity across {\it all} samples of a group while promoting, at the same time, low-density regions amongst data points belonging to different groups. 
Guided by the smoothness assumption that ``similar objects should belong to the same group'', the proposed loss trains the neural network for a classification task, enforcing a consistent labelling amongst samples within a class. We show state-of-the-art results on clustering and image retrieval on several datasets, and show the potential of our method when combined with other techniques such as ensembles. To facilitate further research, we make available the code and the models at \url{https://github.com/dvl-tum/group_loss}.

\end{abstract}

\section{Introduction}
Measuring object similarity is at the core of many important machine learning problems like clustering and object retrieval. 
For visual tasks, this means learning a distance function over images. With the rise of deep neural networks, the focus has rather shifted towards learning a feature embedding that is easily separable using a simple distance function, such as the Euclidean distance. 
In essence, objects of the same class (similar) should be close by in the learned manifold, while objects of a different class (dissimilar) should be far away.

Historically, the best performing approaches get deep feature embeddings from the so-called siamese networks \cite{bromley1994signature}, which are typically trained using the contrastive loss \cite{bromley1994signature} or the triplet loss \cite{DBLP:conf/nips/SchultzJ03,DBLP:journals/jmlr/WeinbergerS09}. 
A clear drawback of these losses is that they only consider pairs or triplets of data points, missing key information about the relationships between all members of the mini-batch. On a mini-batch of size $n$, despite that the number of pairwise relations between samples is $\mathcal{O}(n^2)$, contrastive loss uses only $\mathcal{O}(n/2)$ pairwise relations, while triplet loss uses $\mathcal{O}(2n/3)$ relations.
Additionally, these methods consider only the relations between objects of the same class (positives) and objects of other classes (negatives), without making any distinction that negatives belong to different classes.
This leads to not taking into consideration the global structure of the embedding space, and consequently results in lower clustering and retrieval performance. 
To compensate for that, researchers rely on other tricks to train neural networks for deep metric learning: intelligent sampling \cite{DBLP:conf/iccv/ManmathaWSK17}, multi-task learning \cite{DBLP:conf/cvpr/ZhangZLZ16} or hard-negative mining \cite{DBLP:conf/cvpr/SchroffKP15}. 
Recently, researchers have been increasingly working towards exploiting in a principled way the global structure of the embedding space \cite{DBLP:journals/corr/abs-1906-07589,DBLP:conf/cvpr/Cakir0XKS19,DBLP:conf/cvpr/0003CBS18,DBLP:conf/cvpr/WangHKHGR19}, typically by designing ranking loss functions instead of following the classic triplet formulations.

In a similar spirit, we propose {\it Group Loss}, a novel loss function for deep metric learning that considers the similarity between all samples in a mini-batch. To create the mini-batch, we sample from a fixed number of classes, with samples coming from a class forming a \textit{group}. Thus, each mini-batch consists of several randomly chosen groups, and each group has a fixed number of samples. An iterative, fully-differentiable label propagation algorithm is then used to build feature embeddings which are similar for samples belonging to the same group, and dissimilar otherwise. 

At the core of our method lies an iterative process called {replicator dynamics} \cite{weibull1997evolutionary,DBLP:journals/neco/ErdemP12}, that refines the local information, given by the softmax layer of a neural network, with the global information of the mini-batch given by the similarity between embeddings. 
The driving rationale is that the more similar two samples are, the more they affect each other in choosing their final label and tend to be grouped together in the same group, while dissimilar samples do not affect each other on their choices.
Neural networks optimized with the Group Loss learn to provide similar features for samples belonging to the same class, making clustering and image retrieval easier.

Our \textbf{contribution} in this work is four-fold: 
\begin{itemize}
    \item We propose a novel loss function to train neural networks for deep metric embedding that takes into account the local information of the samples, as well as their similarity.
    \item We propose a differentiable label-propagation iterative model to embed the similarity computation within backpropagation, allowing end-to-end training with our new loss function.
    \item We perform a comprehensive robustness analysis showing the stability of our module with respect to the choice of hyperparameters. 
    \item We show state-of-the-art qualitative and quantitative results in several standard clustering and retrieval datasets. 
\end{itemize}


\begin{figure*}[ht!]
\includegraphics[width=\textwidth, trim={0 4.9cm 0.2cm 0},clip]{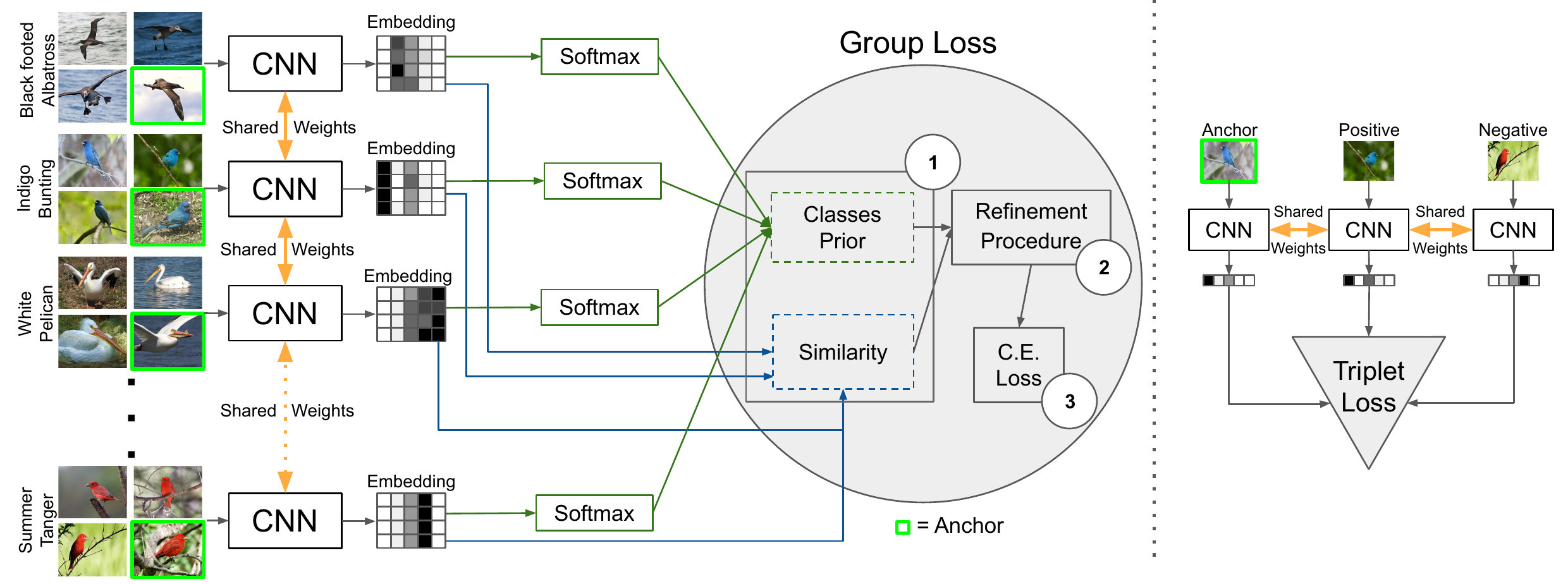} 
\centering
\caption{A comparison between a neural model trained with the Group Loss (left) and the triplet loss (right). Given a mini-batch of images belonging to different classes, their embeddings are computed through a convolutional neural network. Such embeddings are then used to generate a similarity matrix that is fed to the Group Loss along with prior distributions of the images on the possible classes. The green contours around some mini-batch images refer to \emph{anchors}. It is worth noting that, differently from the triplet loss, the Group Loss considers multiple classes and the pairwise relations between all the samples. Numbers from \raisebox{.4pt}{\textcircled{\raisebox{-1pt} {1}}} to \raisebox{.4pt}{\textcircled{\raisebox{-1pt} {3}}} refer to the Group Loss steps, see Sec \ref{sec:gl_overview} for the details.
} 
\label{fig:pipeline}
\end{figure*}

\section{Related Work}

\noindent\textbf{Classical metric learning losses.} The first attempt at using a neural network for feature embedding was done in the seminal work of Siamese Networks \cite{bromley1994signature}. A cost function called \textit{contrastive loss} was designed in such a way as to minimize the distance between pairs of images belonging to the same cluster, and maximize the distance between pairs of images coming from different clusters. In \cite{DBLP:conf/cvpr/ChopraHL05}, researchers used the principle to successfully address the problem of face verification. 
Another line of research on convex approaches for metric learning led to the triplet loss \cite{DBLP:conf/nips/SchultzJ03,DBLP:journals/jmlr/WeinbergerS09}, which was later combined with the expressive power of neural networks \cite{DBLP:conf/cvpr/SchroffKP15}. The main difference from the original Siamese network is that the loss is computed using triplets (an anchor, a positive and a negative data point). 
The loss is defined to make the distance between features of the anchor and the positive sample smaller than the distance between the anchor and the negative sample. The approach was so successful in the field of face recognition and clustering, that soon many works followed. The majority of works on the Siamese architecture consist of finding better cost functions, resulting in better performances on clustering and retrieval. In \cite{DBLP:conf/nips/Sohn16}, the authors generalized the concept of triplet by allowing a joint comparison among $N - 1$ negative examples instead of just one. 
\cite{DBLP:conf/cvpr/SongXJS16} designed an algorithm for taking advantage of the mini-batches during the training process by lifting the vector of pairwise distances within the batch to the matrix of pairwise distances, thus enabling the algorithm to learn feature embedding by optimizing a novel structured prediction objective on the lifted problem. The work was later extended in \cite{DBLP:conf/cvpr/SongJR017}, proposing a new metric learning scheme based on structured prediction that is designed to optimize a clustering quality metric, i.e., the normalized mutual information \cite{DBLP:journals/corr/abs-1110-2515}. Better results were achieved on \cite{DBLP:conf/iccv/WangZWLL17}, where the authors proposed a novel angular loss, which takes angle relationship into account. A very different problem formulation was given by \cite{DBLP:conf/icml/LawUZ17}, where the authors used a spectral clustering-inspired approach to achieve deep embedding. A recent work presents several extensions of the triplet loss that reduce the bias in triplet selection by adaptively correcting the distribution shift on the selected triplets \cite{DBLP:conf/eccv/YuLGDT18}. 

\noindent\textbf{Sampling and ensemble methods.} Knowing that the number of possible triplets is extremely large even for moderately-sized datasets, and having found that the majority of triplets are not informative \cite{DBLP:conf/cvpr/SchroffKP15}, researchers also investigated sampling. In the original triplet loss paper \cite{DBLP:conf/cvpr/SchroffKP15}, it was found that using semi-hard negative mining, the network can be trained to a good performance, but the training is computationally inefficient. 
The work of \cite{DBLP:conf/iccv/ManmathaWSK17} found out that while the majority of research is focused on designing new loss functions, selecting training examples plays an equally important role. The authors proposed a distance-weighted sampling procedure, which selects more informative and stable examples than traditional approaches, achieving excellent results in the process. A similar work was that of \cite{DBLP:conf/eccv/GeHDS18} where the authors proposed a hierarchical version of triplet loss that learns the sampling all-together with the feature embedding. 
The majority of recent works has been focused on complementary research directions such as intelligent sampling \cite{DBLP:conf/iccv/ManmathaWSK17,DBLP:conf/eccv/GeHDS18,DDBLP:conf/cvpr/Duan2019,DDBLP:conf/cvpr/Wand2019,DDBLP:conf/cvpr/Xu2019} or ensemble methods \cite{DBLP:conf/eccv/XuanSP18,DDBLP:conf/cvpr/Sanakoyeu2019,DBLP:conf/eccv/KimGCLK18,DBLP:conf/iccv/OpitzWPB17,DBLP:conf/iccv/YuanYZ17}. As we will show in the experimental section, these can be combined with our novel loss.

\noindent\textbf{Other related problems.} In order to have a focused and concise paper, we mostly discuss methods which tackle image ranking/clustering in standard datasets. Nevertheless, we acknowledge related research on specific applications such as person re-identification or landmark recognition, where researchers are also gravitating towards considering the global structure of the mini-batch. In \cite{DBLP:conf/cvpr/0003CBS18} the authors propose a new hashing method for learning binary embeddings of data by optimizing Average Precision metric. In \cite{DBLP:journals/corr/abs-1906-07589,DBLP:conf/cvpr/0003LS18} authors study novel metric learning functions for local descriptor matching on landmark datasets. \cite{DBLP:conf/cvpr/Cakir0XKS19} designs a novel ranking loss function for the purpose of few-shot learning. Similar works that focus on the global structure have shown impressive results in the field of person re-identification \cite{DBLP:conf/cvpr/ZhaoXC19,DBLP:journals/corr/abs-1904-11397}.

\noindent\textbf{Classification-based losses.} The authors of \cite{DBLP:conf/iccv/Movshovitz-Attias17} proposed to optimize the triplet loss on a different space of triplets than the original samples, consisting of an anchor data point and similar and dissimilar learned proxy data points. These proxies approximate the original data points so that a triplet loss over the proxies is a tight upper bound of the original loss. The final formulation of the loss is shown to be similar to that of softmax cross-entropy loss, challenging the long-hold belief that classification losses are not suitable for the task of metric learning. Recently, the work of \cite{DBLP:journals/corr/abs-1811-12649} showed that a carefully tuned normalized softmax cross-entropy loss function combined with a balanced sampling strategy can achieve competitive results. A similar line of research is that of \cite{DBLP:conf/aaai/ZhengJSZWH19}, where the authors use a combination of normalized-scale layers and Gram-Schmidt optimization to achieve efficient usage of the softmax cross-entropy loss for metric learning. 
The work of \cite{DBLP:journals/corr/abs-1909-05235} goes a step further by taking into consideration the similarity between classes. Furthermore, the authors use multiple centers for class, allowing them to reach state-of-the-art results, at a cost of significantly increasing the number of parameters of the model. In contrast, we propose a novel loss that achieves state-of-the-art results without increasing the number of parameters of the model.

\section{Group Loss}

Most loss functions used for deep metric learning \cite{DBLP:conf/cvpr/SchroffKP15,DBLP:conf/cvpr/SongXJS16,DBLP:conf/nips/Sohn16,DBLP:conf/cvpr/SongJR017,DBLP:conf/iccv/WangZWLL17,DDBLP:conf/cvpr/Wand2019,DBLP:conf/cvpr/WangHKHGR19,DBLP:conf/icml/LawUZ17,DBLP:conf/eccv/GeHDS18,DBLP:conf/iccv/ManmathaWSK17} do not use a classification loss function, e.g., cross-entropy, but rather a loss function based on embedding distances. 
The rationale behind it, is that what matters for a classification network is that the output is correct, which does not necessarily mean that the embeddings of samples belonging to the same class are similar. 
Since each sample is classified independently, it is entirely possible that two images of the same class have two distant embeddings that both allow for a correct classification. 
We argue that a classification loss can still be used for deep metric learning if the decisions do not happen independently for each sample, but rather jointly for a whole {\it group}, i.e., the set of images of the same class in a mini-batch. In this way, the method pushes for images belonging to the same class to have similar embeddings.
%

Towards this end, we propose {\it Group Loss}, an iterative procedure that uses the global information of the mini-batch to refine the local information provided by the softmax layer of a neural network.
This iterative procedure categorizes samples into different \textit{groups}, and enforces consistent labelling among the samples of a group.
While softmax cross-entropy loss judges each sample in isolation, the Group Loss allows us to judge the overall class separation for {\it all} samples.
%
In section \ref{refine}, we show the differences between the softmax cross-entropy loss and Group Loss, and highlight the mathematical properties of our new loss.

\subsection{Overview of Group Loss}\label{sec:gl_overview}

Given a mini-batch $\mathcal{B}$ consisting of $n$ images, consider the problem of assigning a class label $\lambda \in \Lambda = \{1, \dots, m\}$ to each image in $\mathcal{B}$. 
In the remainder of the manuscript, $X=(x_{i\lambda})$ represents a $n \times m$ (non-negative) matrix of image-label soft assignments. In other words, each row of $X$ represents a probability distribution over the label set $\Lambda$ ($\sum_{\lambda} x_{i\lambda}=1 \mbox{ for all } i=1\dots n$). 

Our model consists of the following steps (see also Fig. \ref{fig:pipeline} and Algorithm \ref{algo}):

\begin{enumerate}[label=\protect\circled{\arabic*}]
    \item \textbf{Initialization}: Initialize $X$, the image-label assignment using the softmax outputs of the neural network. Compute the $n \times n$ pairwise similarity matrix $W$ using the neural network embedding.
    \item \textbf{Refinement}: Iteratively, refine $X$ considering the similarities between all the mini-batch images, as encoded in $W$, as well as their labeling preferences.
    \item \textbf{Loss computation}: Compute the cross-entropy loss of the refined probabilities and update the weights of the neural network using backpropagation.
\end{enumerate}

We now provide a more detailed description of the three steps of our method.

\subsection{Initialization}

\noindent{\bf Image-label assignment matrix.} The initial assignment matrix denoted $X(0)$, comes from the softmax output of the neural network. 
We can replace some of the initial assignments in matrix $X$ with one-hot labelings of those samples. We call these randomly chosen samples {\it anchors}, as their assignments do not change during the iterative refine process and consequently do not directly affect the loss function. However, by using their correct label instead of the predicted label (coming from the softmax output of the NN), they guide the remaining samples towards their correct label.

\noindent{\bf Similarity matrix.} A measure of similarity is computed among all pairs of embeddings (computed via a CNN) in $\mathcal{B}$ to generate a similarity matrix $W \in \mathbb{R}^{n \times n}$. 
In this work, we compute the similarity measure using the Pearson's correlation coefficient \cite{DBLP:journals/rsl/Pearson95}:
\begin{equation}\label{eq:pearson}
    \omega(i, j) = \frac{\mathrm{Cov}[\phi(I_i), \phi(I_j)]}{\sqrt{\mathrm{Var}[\phi(I_i)]\mathrm{Var}[\phi(I_j)]}} 
\end{equation} 
for $i\neq j$, and set $\omega(i, i)$ to $0$.
The choice of this measure over other options such as cosine layer, Gaussian kernels, or learned similarities, is motivated by the observation that the correlation coefficient uses data standardization, thus providing invariance to scaling and translation -- unlike the cosine similarity, which is invariant to scaling only -- and it does not require additional hyperparameters, unlike Gaussian kernels \cite{DBLP:conf/icpr/EleziTVP18}. 
The fact that a measure of the linear relationship among features provides a good similarity measure can be explained by the fact that the computed features are actually a highly non-linear function of the inputs. Thus, the linear correlation among the embeddings actually captures a non-linear relationship among the original images.

\subsection{Refinement}
In this core step of the proposed algorithm, the initial assignment matrix $X(0)$ is refined in an iterative manner, taking into account the similarity information provided by matrix $W$. $X$ is updated in accordance with the \emph{smoothness assumption}, which prescribes that similar objects should share the same label.

To this end, let us define the \emph{support} matrix $\Pi=(\pi_{i\lambda}) \in R^{n \times m}$ as
\begin{equation}
    \Pi = W X \label{eqn:pi}
\end{equation}
whose $(i, \lambda)$-component
\begin{equation}
    \pi_{i\lambda} = \sum_{j=1}^{n}w_{ij}x_{j\lambda}
\end{equation}
represents the \emph{support} that the current mini-batch gives to the hypothesis that the $i$-th image in $\mathcal{B}$ belongs to class $\lambda$. Intuitively, in obedience to the smoothness principle, $\pi_{i\lambda}$  is expected to be high if images similar to $i$ are likely to belong to class $\lambda$.

\begin{figure}[t]
    \centering
    \includegraphics[width=0.99\textwidth]{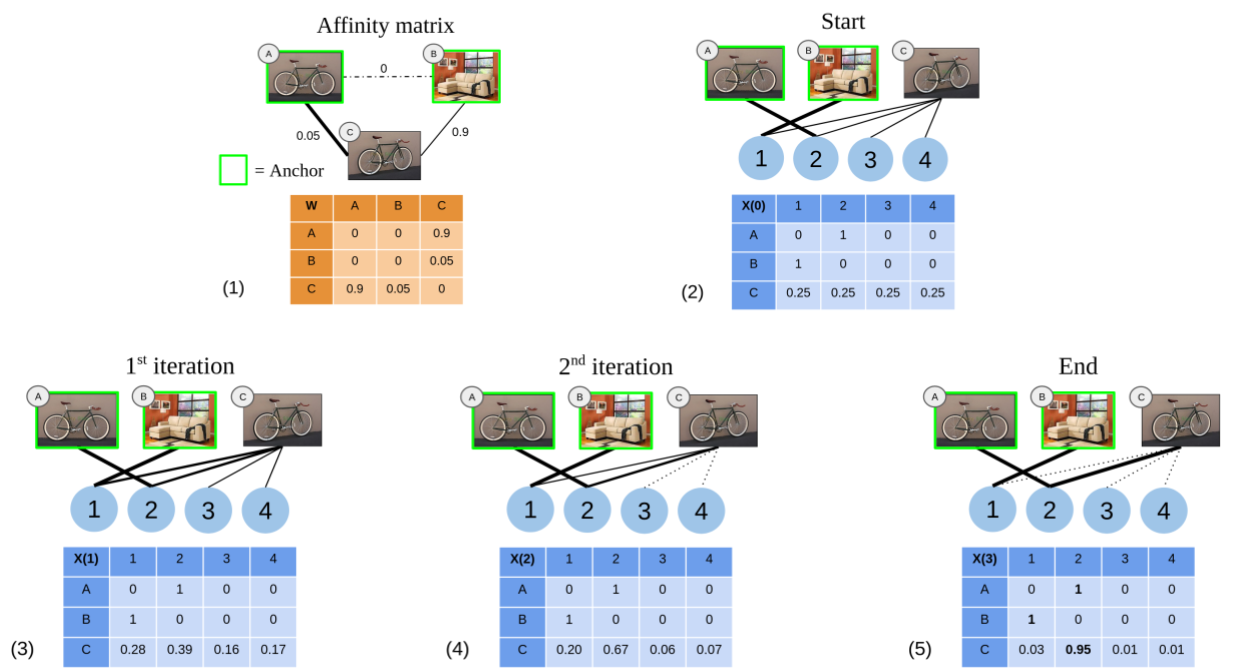}
    \caption{A  toy example of  the  refinement  procedure, where the goal is to classify sample C based on the similarity with samples A and B. (1) The Affinity matrix used to update the soft assignments. (2) The initial labeling of the matrix. (3-4) The process iteratively refines the soft assignment of the unlabeled sample C. (5) At the end of the process, sample C gets the same label of A, (A, C) being more similar than (B, C).}
  \label{fig:gtg_prior}
\end{figure}

Given the initial assignment matrix $X(0)$, our algorithm refines it using the following update rule:
\begin{equation}
x_{i\lambda}(t+1) = \frac{x_{i\lambda}(t)\pi_{i\lambda}(t)}{\sum_{\mu=1}^m{x_{i\mu}(t)\pi_{i\mu}(t)}}
\label{eqn:rd}
\end{equation}
where the denominator represents a normalization factor which guarantees that the rows of the updated matrix sum up to one. 
This is known as multi-population replicator dynamics in evolutionary game theory \cite{weibull1997evolutionary} and is equivalent to nonlinear relaxation labeling processes \cite{RosHumZuc76,DBLP:journals/jmiv/Pelillo97}.

In matrix notation, the update rule (\ref{eqn:rd}) can be written as:
\begin{equation}
X(t+1) = Q^{-1}(t) \left[ X(t) \odot \Pi(t) \right] \label{eq:rd_mat}
\end{equation}
where 
\begin{equation}
Q(t) = \mbox{diag}(\left[X(t) \odot \Pi(t)\right] \mathbb{1} )
\end{equation}
and $\mathbb{1}$ is the all-one $m$-dimensional vector.
$\Pi(t) = W X(t)$ as defined in (\ref{eqn:pi}), and $\odot$ denotes the Hadamard (element-wise) matrix product.
In other words, the diagonal elements of $Q(t)$ represent the normalization factors in (\ref{eqn:rd}), which can also be interpreted as the average support that object $i$ obtains from the current mini-batch at iteration $t$.
Intuitively, the motivation behind our update rule is that at each step of the refinement process, for each image $i$, a label $\lambda$ will increase its probability $x_{i\lambda}$ if and only if its support $\pi_{i\lambda}$ is higher than the average support among all the competing label hypothesis $Q_{ii}$.

Thanks to the Baum-Eagon inequality \cite{DBLP:journals/jmiv/Pelillo97}, it is easy to show that the dynamical system defined by (\ref{eqn:rd}) has very nice convergence properties. In particular, it strictly increases at each step the following functional:

\begin{equation}
F(X) = \sum_{i = 1}^{n}\sum_{j = 1}^{n}\sum_{\lambda = 1}^{m} w_{ij}  x_{i\lambda} x_{j\lambda} \label{eqn:avgconsistency}
\end{equation}
which represents a measure of ``consistency'' of the assignment matrix $X$, in accordance to the smoothness assumption ($F$ rewards assignments where highly similar objects are likely to be assigned the same label).
In other words:
\begin{equation}
F(X(t+1)) \geq F(X(t))
\end{equation}
with equality if and only if $X(t)$ is a stationary point.
Hence, our update rule (\ref{eqn:rd}) is, in fact, an algorithm for maximizing the functional $F$ over the space of row-stochastic matrices.
Note, that this contrasts with classical gradient methods, for which an increase in the objective function is guaranteed only when infinitesimal steps are taken, and determining the optimal step size entails computing higher-order derivatives. Here, instead, the step size is implicit and yet, at each step, the value of the functional increases. 

\label{refine}
\subsection{Loss computation}
Once the labeling assignments converge (or in practice, a maximum number of iterations is reached), we apply the cross-entropy loss to quantify the classification error and backpropagate the gradients. Recall, the refinement procedure is optimized via \textit{replicator dynamics}, as shown in the previous section. By studying Equation (\ref{eq:rd_mat}), it is straightforward to see that it is composed of fully differentiable operations (matrix-vector and scalar products), and so it can be easily integrated within backpropagation. 
Although the refining procedure has no parameters to be learned, its gradients can be backpropagated to the previous layers of the neural network, producing, in turn, better embeddings for similarity computation.

\subsection{Summary of the Group Loss} 
In this section, we proposed the Group Loss function for deep metric learning. 
During training, the Group Loss works by grouping together similar samples based on both the similarity between the samples in the mini-batch and the local information of the samples. The similarity between samples is computed by the correlation between the embeddings obtained from a CNN, while the local information is computed with a softmax layer on the same CNN embeddings. Using an iterative procedure, we combine both sources of information and effectively bring together embeddings of samples that belong to the same class.

During inference, we simply forward pass the images through the neural network
to compute their embeddings, which are directly used for image retrieval within a nearest neighbor search scheme.
The iterative procedure is not used during inference, thus making the feature extraction as fast as that of any other competing method. 

\begin{algorithm}[t] 
\caption{The Group Loss}
\label{algo}
\DontPrintSemicolon 
  \KwInput{input : Set of pre-processed images in the mini-batch $\mathcal{B}$, set of labels $y$, 
  neural network $\phi$ with learnable parameters $\theta$, similarity function $\omega$, number of iterations $T$}  
  
  Compute feature embeddings $\phi(\mathcal{B}, \theta)$ via the forward pass

  Compute the similarity matrix $W = [\omega(i, j)]_{ij}$
  
  Initialize the matrix of priors $X(0)$ from the softmax layer
  
  \For{t = 0, \dots, T-1}
  {
    $Q(t) = \mbox{diag}(\left[X(t) \odot \Pi(t)\right] \mathbb{1} )$\\
    $X(t+1) = Q^{-1}(t) \left[ X(t) \odot \Pi(t) \right]$
  }
      
  Compute the cross-entropy $J(X(T), y)$
  
  Compute the derivatives $\partial J / \partial \theta$ via backpropagation, and update the weights $\theta$
\end{algorithm}

\section{Experiments}
In this section, we compare the Group Loss with state-of-the-art deep metric learning models on both image retrieval and clustering tasks. Our method achieves state-of-the-art results in three public benchmark datasets. 

\subsection{Implementation details}
We use the PyTorch \cite{Paszke17} library for the implementation of the Group Loss. We choose GoogleNet \cite{DBLP:conf/cvpr/SzegedyLJSRAEVR15} with batch-normalization \cite{DBLP:conf/icml/IoffeS15} as the backbone feature extraction network. We pretrain the network on \textit{ILSVRC 2012-CLS} dataset \cite{DBLP:journals/corr/RussakovskyDSKSMHKKBBF14}. For pre-processing, in order to get a fair comparison, we follow the implementation details of \cite{DBLP:conf/cvpr/SongJR017}. The inputs are resized to $256 \times 256$ pixels, and then randomly cropped to $227 \times 227$. Like other methods except for \cite{DBLP:conf/nips/Sohn16}, we use only a center crop during testing time. We train all networks in the classification task for $10$ epochs. We then train the network in the Group Loss task for $60$ epochs using Adam optimizer \cite{DBLP:conf/iclr/Kingma14}. After $30$ epochs, we lower the learning rate by multiplying it by $0.1$. We find the hyperparameters using random search \cite{DBLP:journals/jmlr/BergstraB12}. We use small mini-batches of size $30 - 100$. As sampling strategy, on each mini-batch, we first randomly sample a fixed number of classes, and then for each of the chosen classes, we sample a fixed number of samples.

\subsection{Benchmark datasets}
We perform experiments on $3$ publicly available datasets, evaluating our algorithm on both clustering and retrieval metrics. For training and testing, we follow the conventional splitting procedure \cite{DBLP:conf/cvpr/SongXJS16}.

\textbf{CUB-200-2011} \cite{WahCUB_200_2011} is a dataset containing $200$ species of birds with $11,788$ images, where the first $100$ species ($5,864$ images) are used for training and the remaining $100$ species ($5,924$ images) are used for testing.

\textbf{Cars 196} \cite{KrauseStarkDengFei-Fei_3DRR2013} dataset is composed of $16,185$ images belonging to $196$ classes. We use the first $98$ classes ($8,054$ images) for training and the other $98$ classes ($8,131$ images) for testing.

\textbf{Stanford Online Products} dataset \cite{DBLP:conf/cvpr/SongXJS16}, contains $22,634$ classes with $120,053$ product images in total, where $11,318$ classes ($59,551$ images) are used for training and the remaining $11,316$ classes ($60,502$ images) are used for testing.

\subsection{Evaluation metrics}
Based on the experimental protocol detailed above, we evaluate retrieval performance and clustering quality on data
from unseen classes of the $3$ aforementioned datasets. For the retrieval task, we calculate the percentage of the testing examples whose $K$ nearest neighbors contain at least one example of the same class. This quantity is also known as Recall@K \cite{DBLP:journals/pami/JegouDS11} and is the most used metric for image retrieval evaluation. 

Similar to all other approaches, we perform clustering using K-means algorithm \cite{MacQueen} on the embedded features. Like in other works, we evaluate the clustering quality using the Normalized Mutual Information measure (NMI) \cite{DBLP:journals/corr/abs-1110-2515}. The choice of NMI measure is motivated by the fact that it is invariant to label permutation, a desirable property for cluster evaluation. 

\begin{figure*}[t]
\centering
\begin{minipage}{.333\textwidth}
  \flushleft 
\includegraphics[width=\textwidth, trim={0.5cm 4cm 12cm 0.1cm},clip]{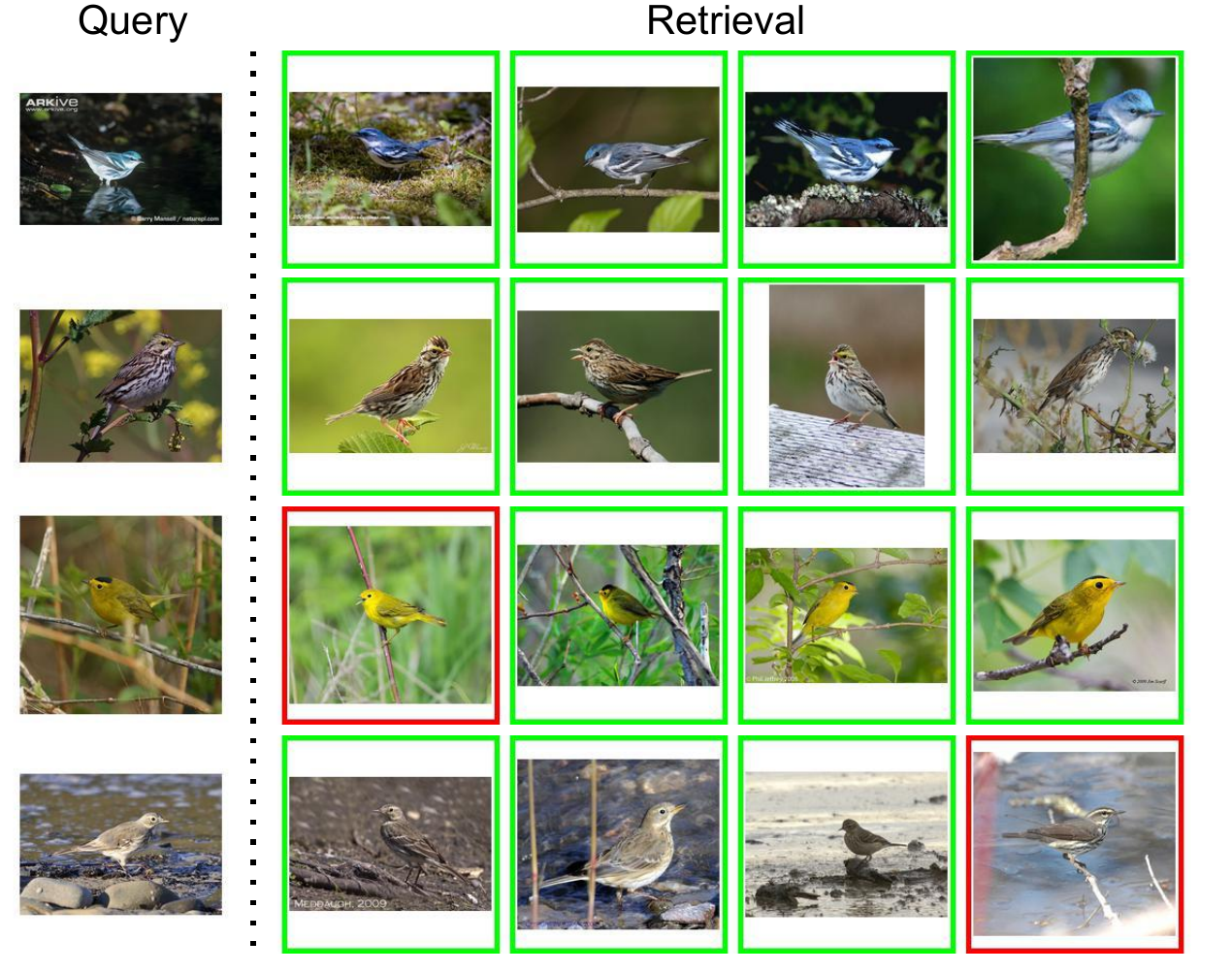}
\end{minipage}%
\begin{minipage}{.333\textwidth}
  \centering
\includegraphics[width=\textwidth, trim={0.5cm 4cm 12cm 0.1cm},clip]{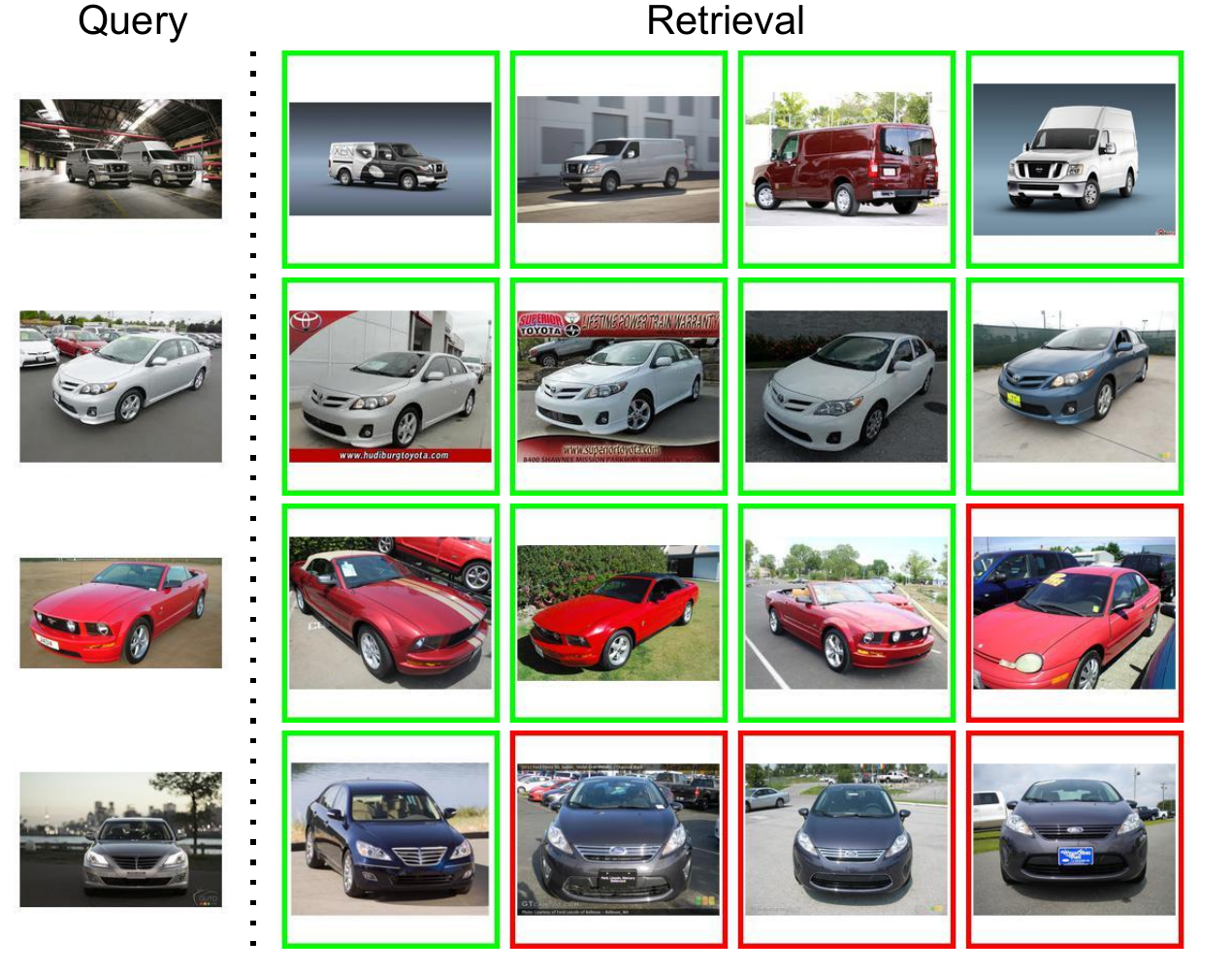}
\end{minipage}%
\begin{minipage}{.333\textwidth}
  \flushright 
\includegraphics[width=\textwidth, trim={0.5cm 4cm 12cm 0.1cm},clip]{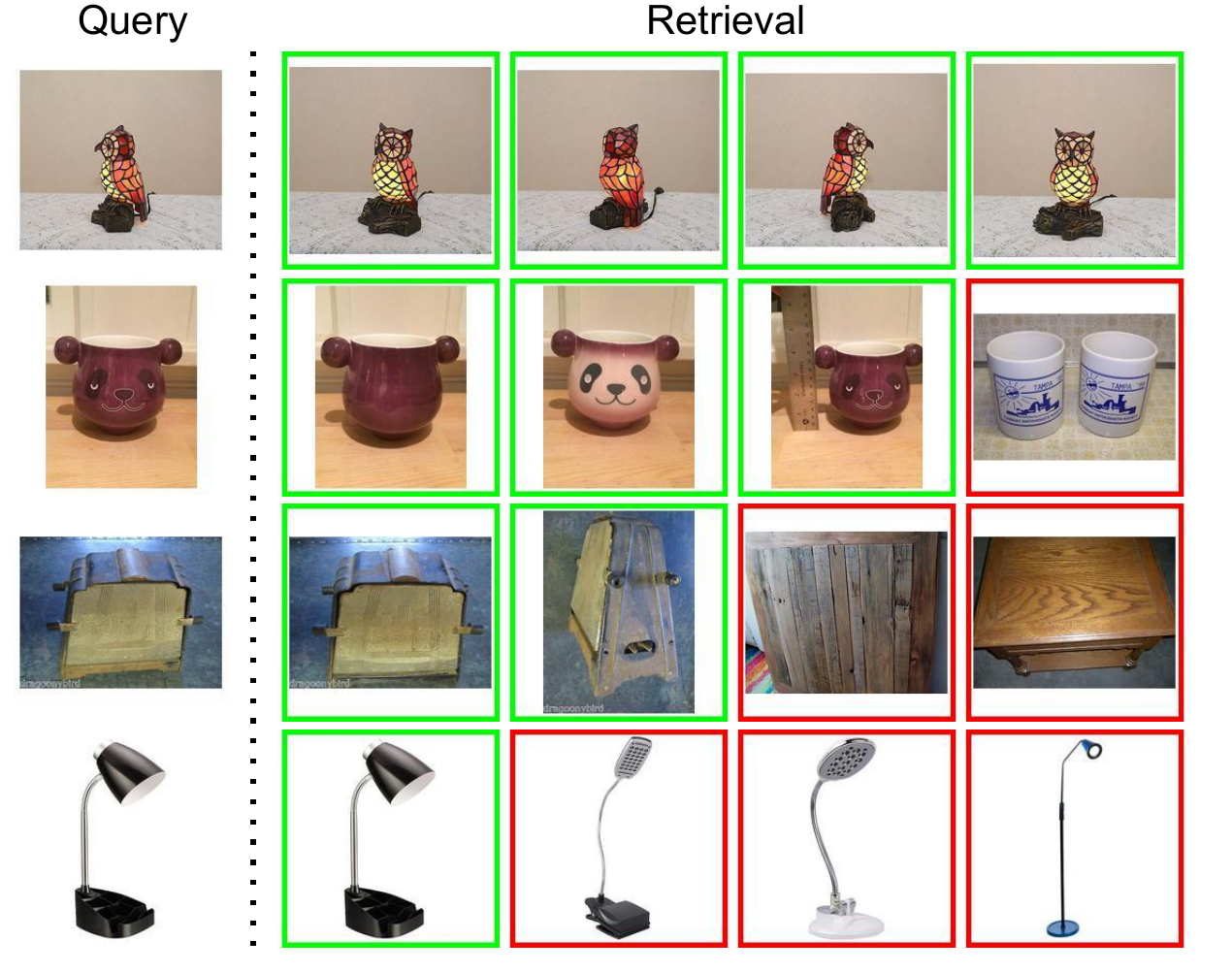}
\end{minipage}

\caption{Retrieval results on a set of images from the
\textit{CUB-200-2011} (left), \textit{Cars 196} (middle), and \textit{Stanford Online Products} (right) datasets using our Group Loss model. The left column contains query images. The results are ranked by distance. The green square indicates that the retrieved image is from the same class as the query image, while the red box indicates that the retrieved image is from a different class.}  
\label{fig:retrieval}
\end{figure*}

\subsection{Results}

We now show the results of our model and comparison to state-of-the-art methods. 
Our main comparison is with other loss functions, e.g., triplet loss. 
 To compare with perpendicular research on intelligent sampling strategies or ensembles, and show the power of the Group Loss, we propose a simple ensemble version of our method. Our ensemble network is built by training $l$ independent neural networks with the same hyperparameter configuration. During inference, their embeddings are concatenated. Note, that this type of ensemble is much simpler than the works of \cite{DBLP:conf/iccv/YuanYZ17,DBLP:conf/eccv/XuanSP18,DBLP:conf/eccv/KimGCLK18,DBLP:journals/corr/abs-1801-04815,DDBLP:conf/cvpr/Sanakoyeu2019}, and is given only to show that, when optimized for performance, our method can be extended to ensembles giving higher clustering and retrieval performance than other methods in the literature. 
%
Finally, in the interest of space, we only present results for Inception network \cite{DBLP:conf/cvpr/SzegedyLJSRAEVR15}, as this is the most popular backbone for the metric learning task, which enables fair comparison among methods. In supplementary material, we present results for other backbones, and include a discussion about the methods that work by increasing the number of parameters (capacity of the network) \cite{DBLP:journals/corr/abs-1909-05235}, or use more expressive network architectures.

\noindent{\textbf{Quantitative results}}

\noindent{\bf Loss comparison.} In Table \ref{tab:res_loss} we present the results of our method and compare them with the results of other approaches. On the \textit{CUB-200-2011} dataset, we outperform the other approaches by a large margin, with the second-best model (Classification \cite{DBLP:journals/corr/abs-1811-12649}) having circa $6$ percentage points($pp$) lower absolute accuracy in Recall@1 metric. On the NMI metric, our method achieves a score of $69.0$ which is  $2.8pp$ higher than the second-best method. Similarly, on \textit{Cars 196}, our method achieves best results on Recall@1, with Classification \cite{DBLP:journals/corr/abs-1811-12649} coming second with a $4pp$ lower score. 
On \textit{Stanford Online Products}, our method reaches the best results on the Recall@1 metric, around $2pp$ higher than Classification \cite{DBLP:journals/corr/abs-1811-12649} and Proxy-NCA \cite{DBLP:conf/iccv/Movshovitz-Attias17}. On the same dataset, when evaluated on the NMI score, our loss outperforms any other method, be those methods that exploit advanced sampling, or ensemble methods.

\begin{table}[t]
\centering
    \resizebox{\textwidth}{!}{%
    \begin{tabular}{@{}l|ccccc|ccccc|cccc@{}}
\toprule

\textbf{} & \multicolumn{5}{c}{CUB-200-2011} & \multicolumn{5}{c}{CARS 196} & \multicolumn{4}{c}{Stanford Online Products}\\ \hline
\textbf{Loss} &  \textbf{R@1} & \textbf{R@2} & \textbf{R@4} & \textbf{R@8} & \textbf{NMI} &  \textbf{R@1} & \textbf{R@2} & \textbf{R@4} & \textbf{R@8} & \textbf{NMI} & \textbf{R@1} & \textbf{R@10} & \textbf{R@100} & \textbf{NMI} \\ \midrule
Triplet \cite{DBLP:conf/cvpr/SchroffKP15} & 42.5 & 55 & 66.4 & 77.2 & 55.3 & 51.5 & 63.8 & 73.5 & 82.4 & 53.4 & 66.7 & 82.4 & 91.9 & 89.5 \\

Lifted Structure \cite{DBLP:conf/cvpr/SongXJS16} & 43.5 & 56.5 & 68.5 & 79.6 & 56.5 &  53.0 & 65.7 & 76.0 & 84.3 & 56.9 & 62.5 & 80.8 & 91.9 & 88.7 \\

Npairs \cite{DBLP:conf/nips/Sohn16} &  51.9 & 64.3 & 74.9 & 83.2 & 60.2 &  68.9 & 78.9 & 85.8 & 90.9 & 62.7 &  66.4 & 82.9 & 92.1 & 87.9  \\

Facility Location \cite{DBLP:conf/cvpr/SongJR017} &  48.1 & 61.4 & 71.8 & 81.9 & 59.2 &  58.1 & 70.6 & 80.3 & 87.8 & 59.0 & 67.0 & 83.7 & 93.2 & 89.5 \\

Angular Loss \cite{DBLP:conf/iccv/WangZWLL17} &  54.7 & 66.3 & 76 & 83.9 & 61.1 &  71.4 & 81.4 & 87.5 & 92.1 & 63.2 &  70.9 & 85.0 & 93.5 & 88.6\\

Proxy-NCA \cite{DBLP:conf/iccv/Movshovitz-Attias17} &  49.2 & 61.9 & 67.9 & 72.4 & 59.5 &  73.2 & 82.4 & 86.4 & 88.7 & 64.9 &  73.7 & - & - & 90.6 \\

Deep Spectral \cite{DBLP:conf/icml/LawUZ17} &  53.2 & 66.1 & 76.7 & 85.2 & 59.2 &  73.1 & 82.2 & 89.0 & 93.0 & 64.3 &  67.6 & 83.7 & 93.3& 89.4 \\ 

Classification \cite{DBLP:journals/corr/abs-1811-12649} &  59.6 & 72 & 81.2 & 88.4 & 66.2 &  81.7 & 88.9 & 93.4 & 96 & 70.5 &  73.8 & 88.1 & \textbf{95} & 89.8 \\

Bias Triplet \cite{DBLP:conf/eccv/YuLGDT18} & 46.6 & 58.6 & 70.0 & - &  - &  79.2 & 86.7 & 91.4  & - & - & 63.0 & 79.8 & 90.7 &- \\ \hline

\textbf{Ours} &  \textbf{65.5} & \textbf{77.0} & \textbf{85.0} & \textbf{91.3} & \textbf{69.0} &  \textbf{85.6} & \textbf{91.2} & \textbf{94.9} & \textbf{97.0} & \textbf{72.7} &  \textbf{75.7} & \textbf{88.2} & 94.8 &\textbf{91.1}\\ \bottomrule 
\end{tabular}}
\caption{Retrieval and Clustering performance on \textit{CUB-200-2011}, \textit{CARS 196} and \textit{Stanford Online Products} datasets. Bold indicates best results. }
\label{tab:res_loss}
\end{table}

\noindent{\bf Loss with ensembles.} In Table \ref{tab:loss_ensembles} we present the results of our ensemble, and compare them with the results of other ensemble and sampling approaches. Our ensemble method (using $5$ neural networks) is the highest performing model in \textit{CUB-200-2011}, outperforming the second-best method (Divide and Conquer \cite{DDBLP:conf/cvpr/Sanakoyeu2019}) by $1pp$ in Recall@1 and by $0.4pp$ in NMI. In \textit{Cars 196} our method outperforms the second best method (ABE 8 \cite{DBLP:conf/eccv/KimGCLK18}) by $2.8pp$ in Recall@1. The second best method in NMI metric is the ensemble version of RLL \cite{DBLP:conf/cvpr/WangHKHGR19} which gets outperformed by $2.4pp$ from the Group Loss. In \textit{Stanford Online Products}, our ensemble reaches the third-highest result on the Recall@1 metric (after RLL \cite{DBLP:conf/cvpr/WangHKHGR19} and GPW \cite{DDBLP:conf/cvpr/Wand2019}) while increasing the gap with the other methods in NMI metric.

\begin{table}[t]
\centering
    \resizebox{\textwidth}{!}{%
    \begin{tabular}{@{}l|ccccc|ccccc|cccc@{}}
\toprule

\textbf{} & \multicolumn{5}{c}{CUB-200-2011} & \multicolumn{5}{c}{CARS 196} & \multicolumn{4}{c}{Stanford Online Products}\\ \hline
\textbf{Loss+Sampling} & \textbf{R@1} & \textbf{R@2} & \textbf{R@4} & \textbf{R@8} & \textbf{NMI} &  \textbf{R@1} & \textbf{R@2} & \textbf{R@4} & \textbf{R@8} & \textbf{NMI} & \textbf{R@1} & \textbf{R@10} & \textbf{R@100} & \textbf{NMI} \\ \midrule

Samp. Matt.  \cite{DBLP:conf/iccv/ManmathaWSK17} &  63.6 & 74.4 & 83.1 & 90.0 & 69.0 &  79.6 & 86.5 & 91.9 & 95.1 & 69.1 &  72.7 & 86.2 & 93.8 & 90.7 \\
Hier. triplet  \cite{DBLP:conf/eccv/GeHDS18} & 57.1 & 68.8 & 78.7 & 86.5 & - &  81.4 & 88.0 & 92.7 & 95.7 &  - & 74.8 & 88.3 & 94.8 & - \\
DAMLRRM  \cite{DDBLP:conf/cvpr/Xu2019} &  55.1 & 66.5 & 76.8 & 85.3 & 61.7 &  73.5 & 82.6 & 89.1 & 93.5 & 64.2 &  69.7 & 85.2 & 93.2 & 88.2\\
DE-DSP  \cite{DDBLP:conf/cvpr/Duan2019} &  53.6 & 65.5 & 76.9 & 61.7 & - &  72.9 & 81.6 & 88.8 & - & 64.4 & 68.9 & 84.0 & 92.6 & 89.2 \\
RLL 1 \cite{DBLP:conf/cvpr/WangHKHGR19} & 57.4 & 69.7 & 79.2 & 86.9 & 63.6 & 74 & 83.6 & 90.1 & 94.1 &   65.4 &   76.1 & 89.1 & 95.4 & 89.7 \\
GPW  \cite{DDBLP:conf/cvpr/Wand2019} & 65.7 & 77.0 & 86.3 & 91.2 & - &  84.1 & 90.4 & 94.0 & 96.5 & - & 78.2 & 90.5 & 96.0 & - \\ \hline \hline

\textbf{Teacher-Student} &  \\ \hline
RKD \cite{DBLP:conf/cvpr/Park2019} &  61.4 & 73.0 & 81.9 & 89.0 & - &  82.3 & 89.8 & 94.2 & 96.6 & - & 75.1 & 88.3 & 95.2  & -\\ \hline \hline

\textbf{Loss+Ensembles} &  \\ \hline
BIER 6  \cite{DBLP:conf/iccv/OpitzWPB17} & 55.3 & 67.2 & 76.9 & 85.1 &  - & 75.0 & 83.9 & 90.3 & 94.3 &  - &  72.7 & 86.5 & 94.0  & -\\
HDC 3  \cite{DBLP:conf/iccv/YuanYZ17} &  54.6 & 66.8 & 77.6 & 85.9 & - &  78.0 & 85.8 & 91.1 & 95.1 & - & 70.1 & 84.9 & 93.2 & - \\
ABE 2  \cite{DBLP:conf/eccv/KimGCLK18} &  55.7 & 67.9 & 78.3 & 85.5 & - & 76.8 & 84.9 & 90.2 & 94.0 & - & 75.4 & 88.0 & 94.7  & -  \\
ABE 8  \cite{DBLP:conf/eccv/KimGCLK18} &  60.6 & 71.5 & 79.8 & 87.4 & - & 85.2 & 90.5 & 94.0 & 96.1 &  - & 76.3 & 88.4 &  94.8 & - \\
A-BIER 6 \cite{DBLP:journals/corr/abs-1801-04815} & 57.5 & 68.7 & 78.3 & 86.2 & - &  82.0 & 89.0 & 93.2 & 96.1 &  - & 74.2 & 86.9 & 94.0 & - \\
D and C 8  \cite{DDBLP:conf/cvpr/Sanakoyeu2019} & 65.9 & 76.6 & 84.4 & 90.6 & 69.6 & 84.6 & 90.7 & 94.1 & 96.5 &  70.3 &   75.9   & 88.4 & 94.9 &90.2 \\ 
RLL 3  \cite{DBLP:conf/cvpr/WangHKHGR19} & 61.3 & 72.7 & 82.7 & 89.4 & 66.1 & 82.1 & 89.3 & 93.7 & 96.7 & 71.8 &  \textbf{79.8} & \textbf{91.3} & \textbf{96.3} & 90.4\\
\hline
\textbf{Ours 2-ensemble} & 65.8 & 76.7 & 85.2 & 91.2 & 68.5 &  86.2 & 91.6 & 95.0 & 97.1 & \textbf{91.1} & 75.9 & 88.0 & 94.5 & 72.6\\
\textbf{Ours 5-ensemble} &  \textbf{66.9} & \textbf{77.1} & \textbf{85.4} & \textbf{91.5} & \textbf{70.0} &  \textbf{88.0} & \textbf{92.5} & \textbf{95.7} & \textbf{97.5} & \textbf{74.2} &  76.3 & 88.3 & 94.6 & \textbf{91.1} \\ \bottomrule
\end{tabular}}
\caption{Retrieval and Clustering performance of our ensemble compared with other ensemble and sampling methods. Bold indicates best results. }
\label{tab:loss_ensembles}
\end{table}

\noindent{\textbf{Qualitative results}}

In Fig. \ref{fig:retrieval} we present qualitative results on the retrieval task in all three datasets. In all cases, the query image is given on the left, with the four nearest neighbors given on the right. 
Green boxes indicate the cases where the retrieved image is of the same class as the query image, and red boxes indicate a different class. 
As we can see, our model is able to perform well even in cases where the images suffer from occlusion and rotation. On the \textit{Cars 196} dataset, we see a successful retrieval even when the query image is taken indoors and the retrieved image outdoors, and vice-versa. 
The first example of \textit{Cars 196} dataset is of particular interest. Despite that the query image contains $2$ cars, its four nearest neighbors have the same class as the query image, showing the robustness of the algorithm to uncommon input image configurations. We provide the results of t-SNE \cite{DBLP:journals/ml/MaatenH12} projection in the supplementary material.

\subsection{Robustness analysis}

\begin{figure*}[t]
\centering
\begin{minipage}[t]{.32\textwidth}
  \centering
\includegraphics[width=\textwidth, trim={0cm 0cm 0cm 0cm},clip]{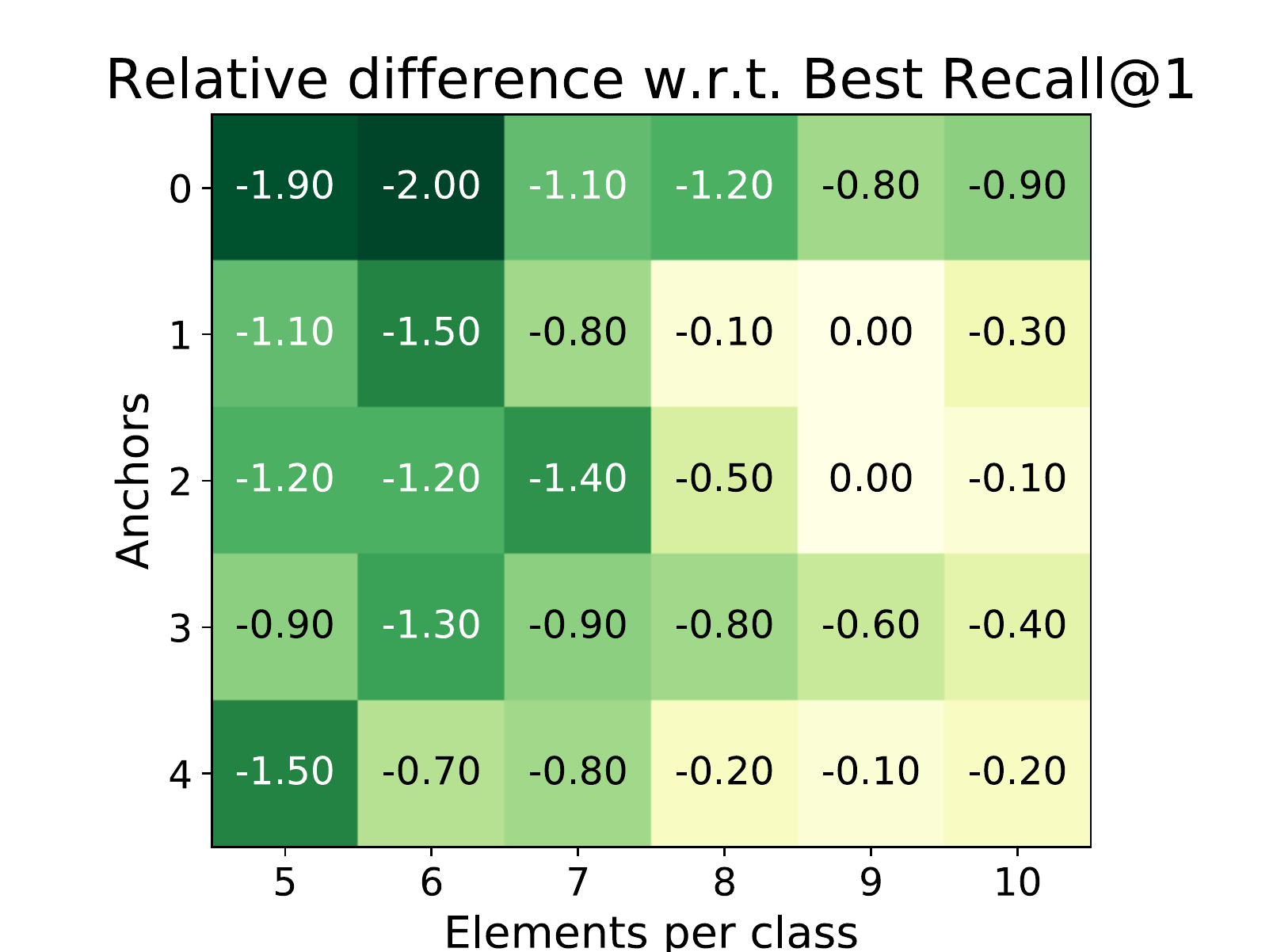} 
  \caption{The effect of the number of anchors and the number of samples per class.}
  \label{fig:abl_labeledpoints}
\end{minipage}
\hfill %
\begin{minipage}[t]{.32\textwidth}
  \centering
\includegraphics[width=\textwidth, trim={0cm 0cm 0cm 0cm},clip]{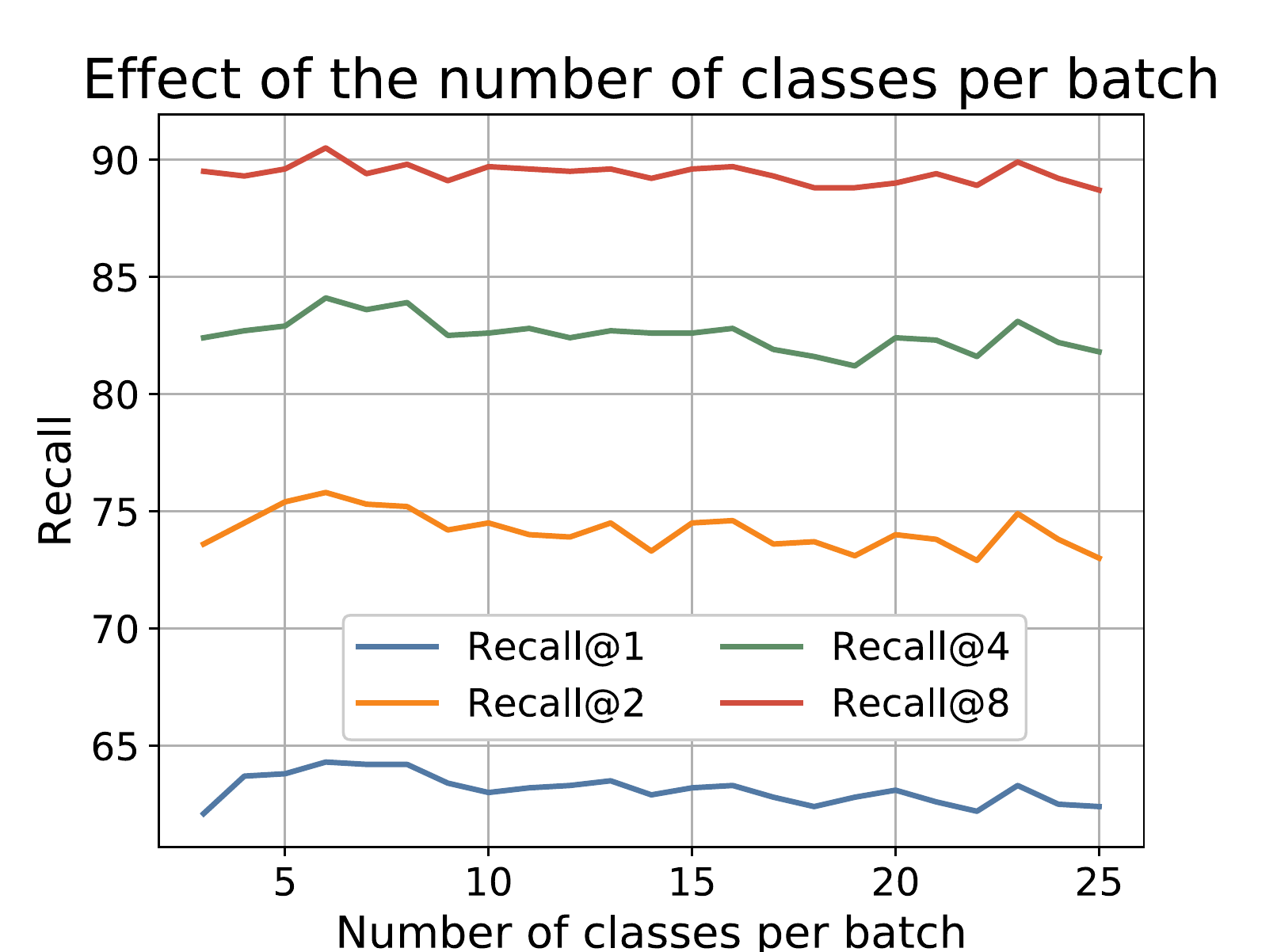} 
  \caption{The effect of the number of classes per mini-batch.}
  \label{fig:abl_classes}
\end{minipage}
\hfill %
\begin{minipage}[t]{.32\textwidth}
  \centering
\includegraphics[width=\textwidth, trim={0cm 0.4cm 0cm 0cm},clip]{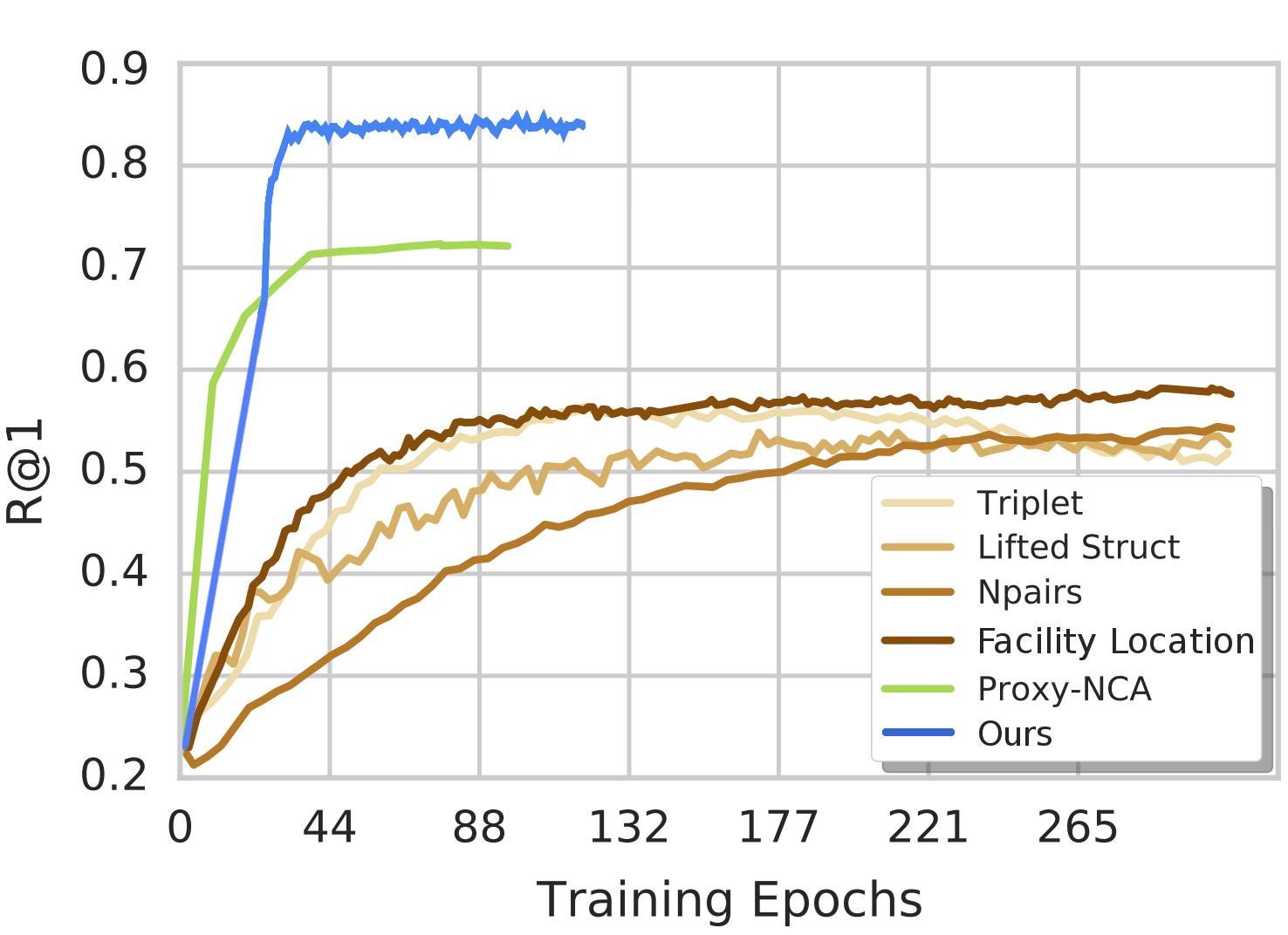} 
  \caption{Recall@1 as a function of training epochs on Cars196 dataset. Figure adapted from \cite{DBLP:conf/iccv/Movshovitz-Attias17}.}
  \label{fig:abl_convergence}
\end{minipage}
\end{figure*}

\textbf{Number of anchors.} 
In Fig.~\ref{fig:abl_labeledpoints}, we show the effect of the number of anchors with respect to the number of samples per class. We do the analysis on \textit{CUB-200-2011} dataset and give a similar analysis for \textit{CARS} dataset in the supplementary material. The results reported are the percentage point differences in terms of Recall@1 with respect to the best performing set of parameters (see $Recall@1=64.3$ in Tab.~\ref{tab:res_loss}). 
The number of anchors ranges from 0 to 4, while the number of samples per class varies from 5 to 10. 
It is worth noting that our best setting considers 1 or 2 anchors over 9 samples. Moreover, even when we do not use any anchor, the difference in Recall@1 is no more than $2pp$.

\textbf{Number of classes per mini-batch.}
In Fig.~\ref{fig:abl_classes}, we present the change in Recall@1 on the \textit{CUB-200-2011} dataset if we increase the number of classes we sample at each iteration. The best results are reached when the number of classes is not too large. This is a welcome property, as we are able to train on small mini-batches, known to achieve better generalization performance \cite{DBLP:conf/iclr/KeskarMNST17}.

\textbf{Convergence rate.} In Fig.~\ref{fig:abl_convergence}, we present the convergence rate of the model on the \textit{Cars 196} dataset. Within the first $30$ epochs, our model achieves state-of-the-art results, making our model significantly faster than other approaches. The other models except Proxy-NCA \cite{DBLP:conf/iccv/Movshovitz-Attias17}, need hundreds of epochs to converge. 
%

\begin{figure*}[t]
\centering
\begin{minipage}[t]{.49\textwidth}
  \centering
\includegraphics[width=\textwidth, trim={0cm 0cm 0cm 0cm},clip]{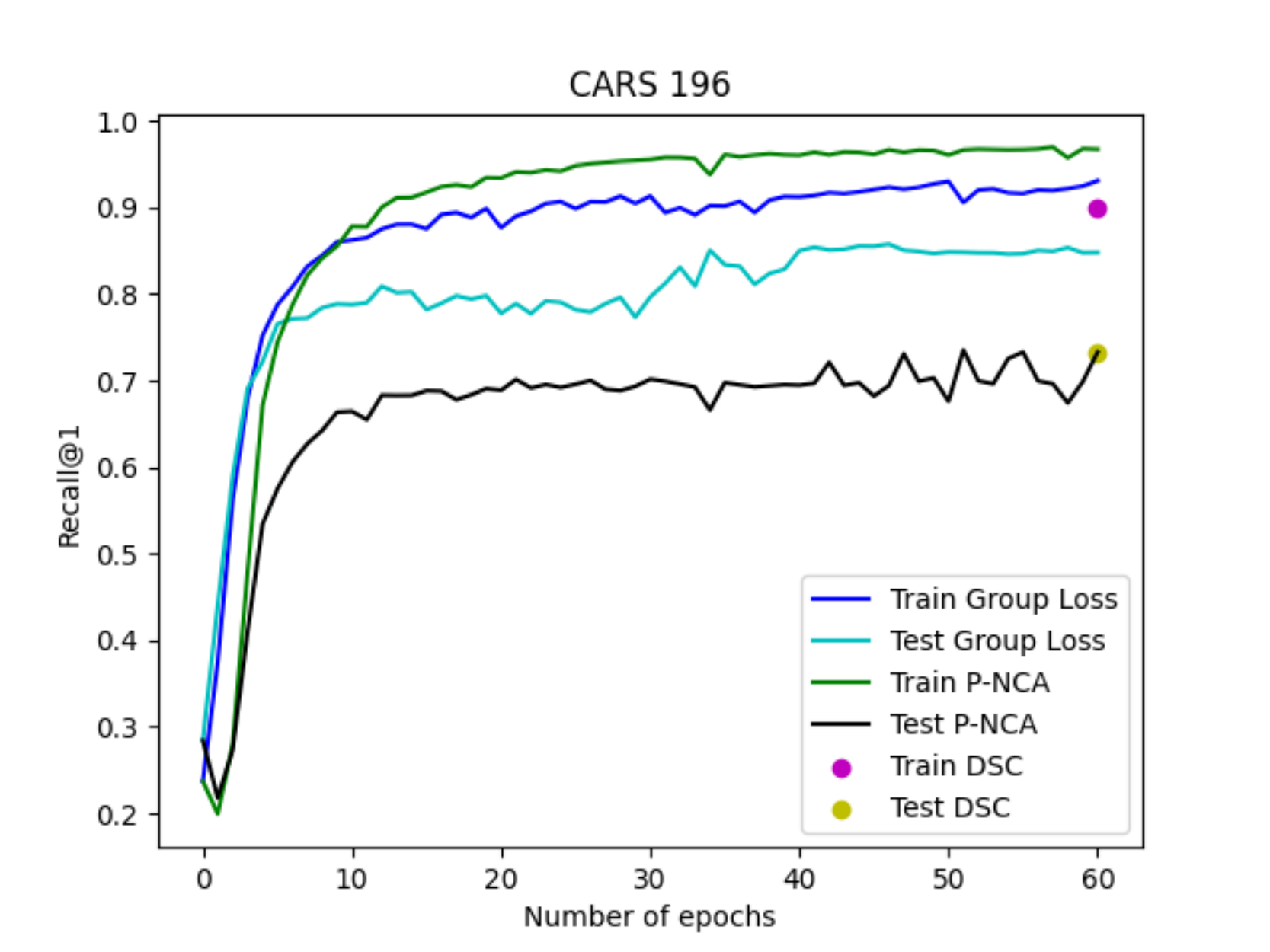} 
  \caption{Training vs testing Recall@1 curves on \textit{Cars 196} dataset.}
  \label{fig:regularization_cars}
\end{minipage}
\hfill %
\begin{minipage}[t]{.49\textwidth}
  \centering
\includegraphics[width=\textwidth, trim={0cm 0cm 0cm 0cm},clip]{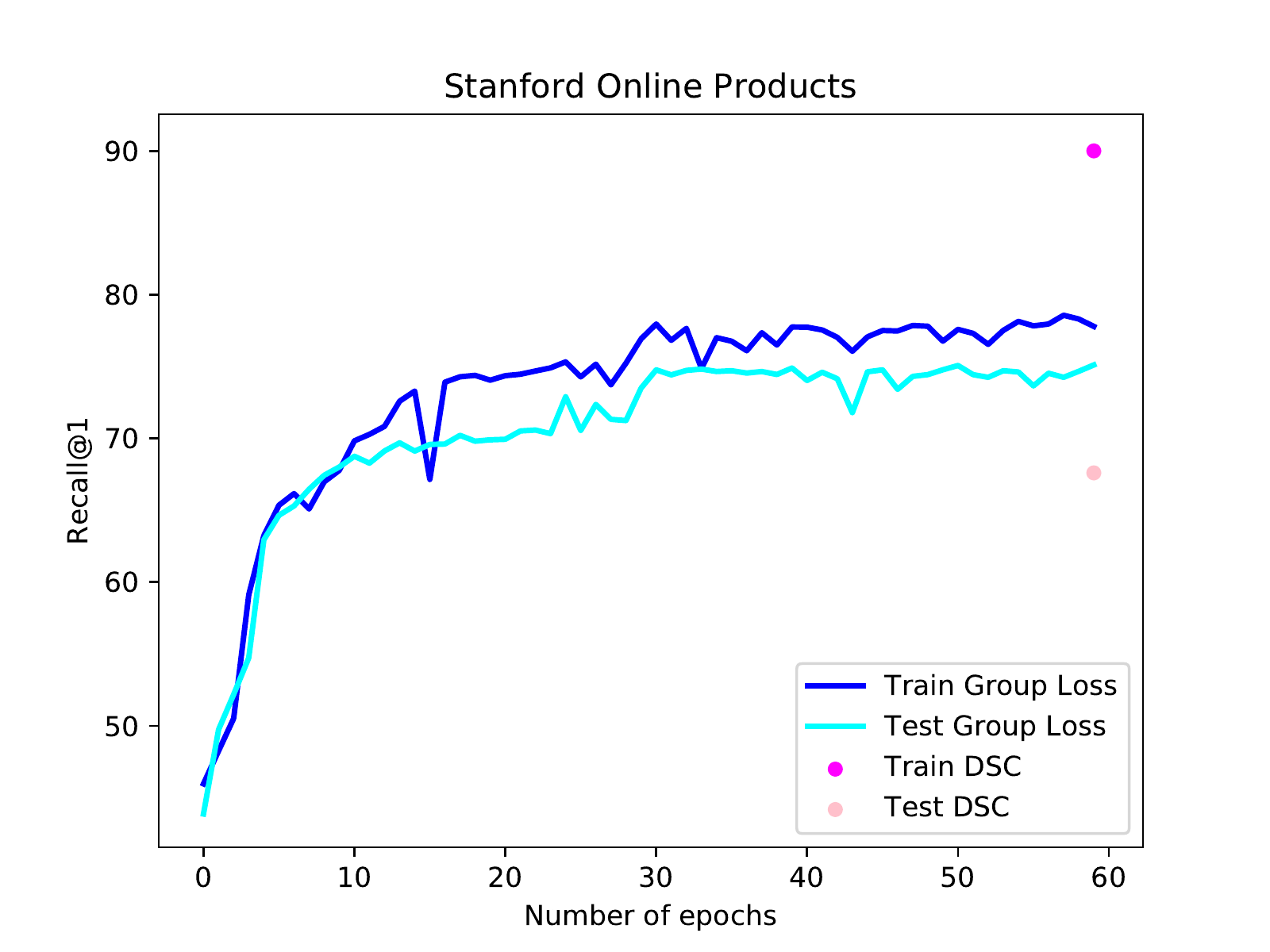} 
  \caption{Training vs testing Recall@1 curves on \textit{Stanford Online Products} dataset.}
  \label{fig:regularization_sop}
\end{minipage}
\hfill %
\end{figure*}

\textbf{Implicit regularization and less overfitting.} In Figures \ref{fig:regularization_cars} and \ref{fig:regularization_sop}, we compare the results of training vs. testing on \textit{Cars 196} \cite{KrauseStarkDengFei-Fei_3DRR2013} and \textit{Stanford Online Products} \cite{DBLP:conf/cvpr/SongXJS16} datasets.
We see that the difference between Recall@1 at train and test time is small, especially on \textit{Stanford Online Products} dataset.
On \textit{Cars 196} the best results we get for the training set are circa $93\%$ in the Recall@1 measure, only $7.5$ percentage points ($pp$) better than what we reach in the testing set.
From the works we compared the results with, the only one which reports the results on the training set is 
\cite{DBLP:conf/icml/LawUZ17}. They reported results of over $90\%$ in 
all three datasets (for the training sets), much above the test set accuracy which lies at $73.1\%$ on \textit{Cars 196} and $67.6\%$ on \textit{Stanford Online Products} dataset. \cite{DBLP:conf/wacv/VoH19} also provides results, but it uses a different network.
%

We further implement the P-NCA \cite{DBLP:conf/iccv/Movshovitz-Attias17} loss function and perform a similar experiment, in order to be able to compare training and test accuracies directly with our method. In Figure \ref{fig:regularization_cars}, we show the training and testing curves of P-NCA on the \textit{Cars 196} \cite{KrauseStarkDengFei-Fei_3DRR2013} dataset. We see that while in the training set, P-NCA reaches results of $3pp$ higher than our method, in the testing set, our method outperforms P-NCA by around $10pp$. Unfortunately, we were unable to reproduce the results of the paper \cite{DBLP:conf/iccv/Movshovitz-Attias17} on \textit{Stanford Online Products} dataset. Furthermore, even when we turn off $L2$-regularization, the generalization performance of our method does not drop at all.
Our intuition is that by taking into account the structure of the entire manifold of the dataset, our method introduces a form of regularization.
We can clearly see a smaller gap between training and test results when compared to competing methods, indicating less overfitting.
%

\section{Conclusions and Future Work}

In this work, we propose the Group Loss, a novel loss function for metric learning. By considering the content of a mini-batch, it promotes embedding similarity across all samples of
the same class, while enforcing dissimilarity for elements of different classes. 
This is achieved with a differentiable layer that is used to train a convolutional network in an end-to-end fashion.
Our model outperforms state-of-the-art methods on several datasets, and shows fast convergence.
In our work, we did not consider any advanced sampling strategy. Instead, we randomly sample objects from a few classes at each iteration. Sampling has shown to have a very important role in feature embedding \cite{DBLP:conf/iccv/ManmathaWSK17}. As future work, we will explore sampling techniques which can be suitable for our module. 

\small \textbf{Acknowledgements.} This research was partially funded by the Humboldt Foundation through the Sofja Kovalevskaja Award. We thank Michele Fenzi, Maxim Maximov and Guillem Braso Andilla for useful discussions.

\appendix

\section{Robustness Analysis of CARS 196 Dataset}
In the main work, we showed the robustness analysis on the \textit{CUB-200-2011} \cite{WahCUB_200_2011} dataset (see Figure 4 in the main paper). Here, we report the same analysis for the \textit{Cars 196} \cite{KrauseStarkDengFei-Fei_3DRR2013} dataset. This leads us to the same conclusions as shown in the main paper.

\begin{figure}
  \centering
  \includegraphics[width=0.75\textwidth]{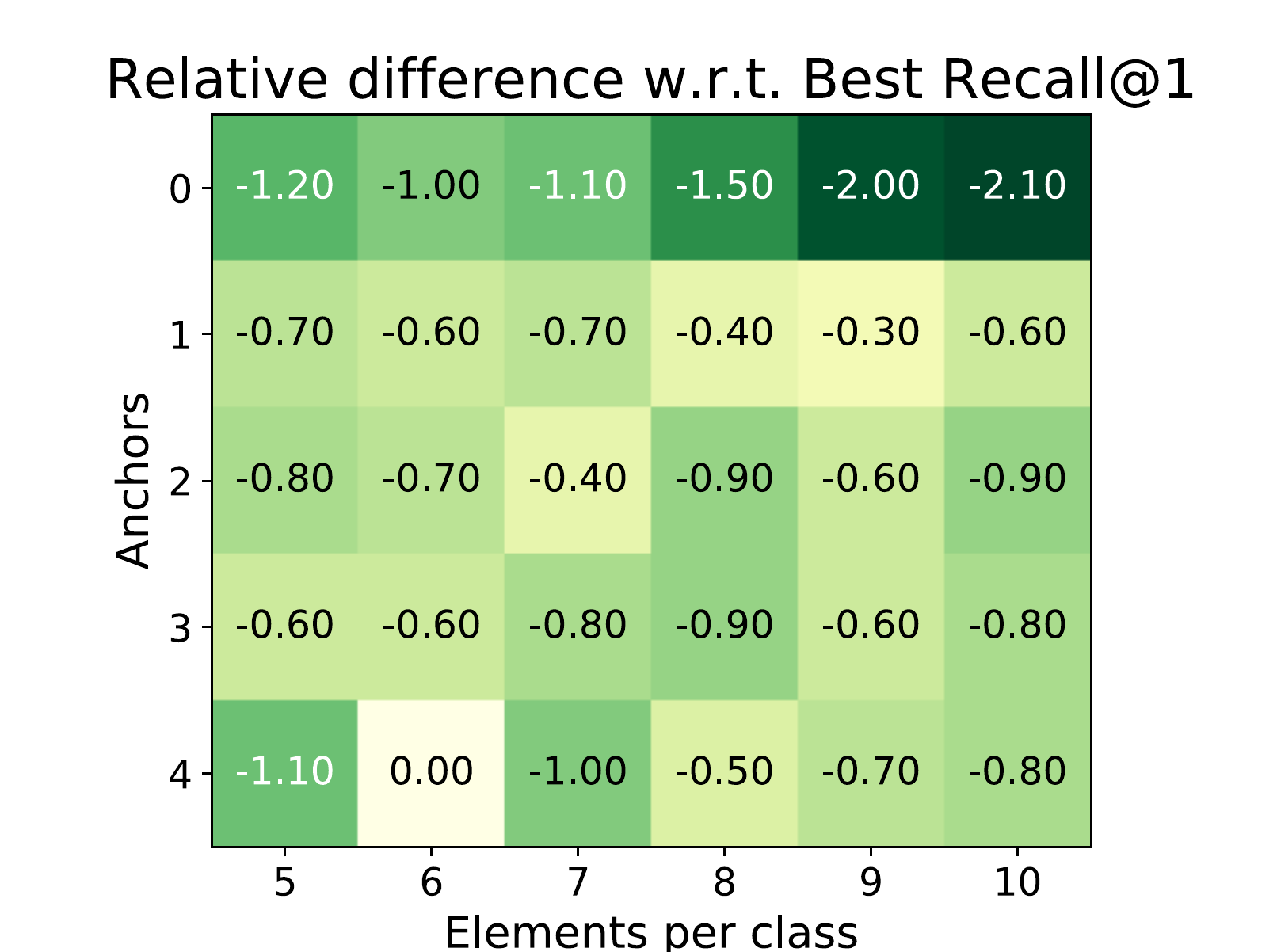}
  \caption{The effect of the number of anchors and the number of samples per class.}
  \label{fig:cars_anchors}
\end{figure}

We do a grid search over the total number of elements per class versus the number of anchors, as we did for the experiment in the main paper. We increase the number of elements per class from $5$ to $10$, and in each case, we vary the number of anchors from $0$ to $4$. We show the results in Fig. \ref{fig:cars_anchors}.
Note, the results decrease mainly when we do not have any labeled sample, i.e., when we use zero anchors.
The method shows the same robustness as on the \textit{CUB-200-2011} \cite{WahCUB_200_2011} dataset, with the best result being only $2.1$ percentage points better at the Recall@1 metric than the worst result.

\section{More Implementation Details}
We first pre-train all networks in the classification task for $10$ epochs. We then train our networks on all three datasets \cite{WahCUB_200_2011,KrauseStarkDengFei-Fei_3DRR2013,DBLP:conf/cvpr/SongXJS16} for $60$ epochs. During training, we use a simple learning rate scheduling in which we divide the learning rate by $10$ after the first $30$ epochs.

We find all hyperparameters using random search \cite{DBLP:journals/jmlr/BergstraB12}. For the weight decay ($L2$-regularization) parameter, we search over the interval $[0.1, 10^{-16}]$, while for learning rate we search over the interval $[0.1, 10^{-5}]$, choosing $0.0002$ as the learning rate for all networks and all datasets. During inference we normalize the features using the L2-norm.

We achieve the best results with a regularization parameter set to $10^{-6}$ for \textit{CUB-200-2011}, $10^{-7}$ for \textit{Cars 196} dataset, and $10^{-12}$ for \textit{Stanford Online Products} dataset. This further strengthens our intuition that the method is implicitly regularized and it does not require strong regularization. 

\section{Temperature Scaling}
We mentioned in the main paper that as input to the Group Loss (step 3 of the algorithm) we initialize the matrix of priors $X(0)$ from the softmax layer of the neural network. 
Following the works of \cite{DBLP:journals/corr/HintonVD15,DBLP:conf/icml/GuoPSW17,DBLP:conf/nips/BerthelotCGPOR19,DBLP:journals/corr/abs-1811-12649}, we apply a sharpening function to reduce the entropy of the softmax distribution. We use the common approach of adjusting the {\it temperature} of this categorical distribution, known as temperature scaling. 
Intuitively, this procedure calibrates our network and in turn, provides more informative prior to the dynamical system. Additionally, this calibration allows the dynamical system to be more effective in adjusting the predictions, i.e, it is easier to change the probability of a class if its initial value is $0.6$ rather than $0.95$. The function is implemented using the following equation:

\begin{equation}
    T_{softmax}(z_i) = \frac{e^{z_i/T}}{\sum_i{e^{z_i/T}}},
\end{equation}

which can be efficiently implemented by simply dividing the prediction logits by a constant $T$.

Recent works in supervised learning \cite{DBLP:conf/icml/GuoPSW17} and semi-supervised learning \cite{DBLP:conf/nips/BerthelotCGPOR19} have found that temperature calibration improves the accuracy for the image classification task. We arrive at similar conclusions for the task of metric learning, obtaining $2.5pp$ better Recall@1 scores on \textit{CUB-200-2011} \cite{WahCUB_200_2011} and $2pp$ better scores on \textit{Cars 196} \cite{KrauseStarkDengFei-Fei_3DRR2013}.
Note, the methods of Table 1 (main paper) that use a classification loss, use also temperature scaling.

\section{Other Backbones}

In the main paper, we perform all experiments using a GoogleNet backbone with batch normalization. This choice is motivated by the fact that most methods use this backbone, making comparisons fair. 
In this section, we explore the performance of our method for other backbone architectures, to show the generality of our proposed loss formulation. 
We choose to train a few networks from Densenet family \cite{DBLP:conf/cvpr/HuangLMW17}. Densenets are a modern CNN architecture which show similar classification accuracy to GoogleNet in most tasks (so they are a similarly strong classification baseline \footnote{The classification accuracy of different backbones can be found in the following link: \url{https://pytorch.org/docs/stable/torchvision/models.html}. BN-Inception's top 1/top 5 error is 7.8\%/25.2\%, very similar to those of Densenet121 (7.8\%/25.4\%).}). 
Furthermore, by training multiple networks of the same family, we can study the effect of the capacity of the network, i.e., how much can we gain from using a larger network? 
Finally, we are interested in studying if the choice of hyperparameters can be transferred from one backbone to another.

We present the results of our method using Densenet backbones in Tab. \ref{tab:densenets}. 
We use the same hyperparameters as the ones used for the GoogleNet experiments, reaching state-of-the-art results on both \textit{CARS 196} \cite{KrauseStarkDengFei-Fei_3DRR2013} and \textit{Stanford Online Products} \cite{DBLP:conf/cvpr/SongXJS16} datasets, even compared to ensemble and sampling methods. 
The results in \textit{Stanford Online Products} \cite{DBLP:conf/cvpr/SongXJS16} are particularly impressive considering that this is the first time any method in the literature has broken the $80$ point barrier in Recall@1 metric. 
We also reach state-of-the-art results on the \textit{CUB-200-2011} \cite{WahCUB_200_2011} dataset when we consider only methods that do not use ensembles (with the Group Loss ensemble reaching the highest results in this dataset). 
We observe a clear trend when increasing the number of parameters (weights), with the best results on both \textit{CARS 196} \cite{KrauseStarkDengFei-Fei_3DRR2013} and \textit{Stanford Online Products} \cite{DBLP:conf/cvpr/SongXJS16} datasets being achieved by the largest network, Densenet161 (whom has a lower number of convolutional layers than Densenet169 and Densenet201, but it has a higher number of weights/parameters).

Finally, we study the effects of hyperparameter optimization. Despite that the networks reached state-of-the-art results even without any hyperparameter tuning, we expect a minimum amount of hyperparameters tuning to help. 
To this end, we used random search \cite{DBLP:journals/jmlr/BergstraB12} to optimize the hyperparameters of our best network on the \textit{CARS 196} \cite{KrauseStarkDengFei-Fei_3DRR2013} dataset. We reach a $90.7$ score ($2pp$ higher score than the network with default hyperparameters) in Recall@1, and $77.6$ score ($3pp$ higher score than the network with default hyperparameters) in NMI metric, showing that individual hyperparameter optimization can boost the performance. 
The score of $90.7$ in Recall@1 is not only by far the highest score ever achieved, but also the first time any method has broken the $90$ point barrier in Recall@1 metric when evaluated on the \textit{CARS 196} \cite{KrauseStarkDengFei-Fei_3DRR2013} dataset.

\begin{table}[t]
\centering
    \resizebox{\textwidth}{!}{
    \begin{tabular}{@{}l||l|l|l||l|l|l||l|l|l@{}}
\toprule
\multicolumn{1}{c}{Model} & \multicolumn{3}{c}{CUB} & \multicolumn{3}{c}{CARS} & \multicolumn{3}{c}{SOP}  \\ 
\hline
\multicolumn{1}{c}{}                        & \multicolumn{1}{c}{Params} & \multicolumn{1}{c}{R@1} & \multicolumn{1}{c}{NMI} & \multicolumn{1}{c}{Params} & \multicolumn{1}{c}{R@1} & \multicolumn{1}{c}{NMI} & \multicolumn{1}{c}{Params} & \multicolumn{1}{c}{R@1} & \multicolumn{1}{c}{NMI}\\ \hline
GL Densenet121                 & 7056356      &   \textbf{65.5}  & 69.4 &  7054306    & 88.1  & 74.2  &  18554806   &   78.2 &   91.5     \\
GL Densenet161                 &  26692900     &  64.7  &  68.7 &  26688482    & \textbf{88.7}  & 74.6 & 51473462    & \textbf{80.3} &  \textbf{92.3}        \\
GL Densenet169                 &  12650980     &  65.4   & \textbf{69.5} &  12647650    & 88.4 & 75.2 & 31328950    &    79.4 &  92.0    \\
GL Densenet201                 &  18285028     &  63.7  & 68.4  &  18281186    &  88.6 & \textbf{75.8}  & 39834806    &    79.8   &  92.1     \\ \hline
GL Inception v2                 &   10845216    &  \textbf{65.5}  & 69.0  &  10846240    & 85.6   & 72.7 & 16589856    & 75.7   & 91.1       \\ \hline \hline
SofTriple \cite{DBLP:journals/corr/abs-1909-05235}             & 11307040      &  65.4  & 69.3  &   11296800   & 84.5  & 70.1 &  68743200   & 78.3  & 92        \\
\hline \bottomrule 

\end{tabular}}
\caption{The results of Group Loss in Densenet backbones and comparisons with SoftTriple loss \cite{DBLP:journals/corr/abs-1909-05235}}
\label{tab:densenets}
\end{table}

\section{Comparisons with SoftTriple Loss \cite{DBLP:journals/corr/abs-1909-05235}}

A recent paper (SoftTriple loss \cite{DBLP:journals/corr/abs-1909-05235}, ICCV 2019) explores another type of classification loss for the problem of metric learning. The main difference between our method and \cite{DBLP:journals/corr/abs-1909-05235} is that our method checks the similarity between samples, and then refines the predicted probabilities (via a dynamical system) based on that information. \cite{DBLP:journals/corr/abs-1909-05235} instead deals with the intra-class variability, but does not explicitly take into account the similarity between the samples in the mini-batch. 
They propose to add a new layer with $10$ units per class. 

We compare the results of \cite{DBLP:journals/corr/abs-1909-05235} with our method in Tab. \ref{tab:densenets}. SoftTriple loss \cite{DBLP:journals/corr/abs-1909-05235} reaches a higher result than our method in all three datasets in Recall@1 metric, and higher results than the Group Loss on the \textit{CUB-200-2011} and \textit{Stanford Online Products} datasets in NMI metric. However, this comes at a cost of significantly increasing the number of parameters. 
On the \textit{Stanford Online Products} dataset in particular, the number of parameters of \cite{DBLP:journals/corr/abs-1909-05235} is $68.7$ million. In comparison, we (and the other methods we compare the results with in the main paper) use only $16.6$ million parameters. 
In effect, their increase in performance comes at the cost of using a neural network which is 4 times larger as ours, making results not directly comparable. 
Furthermore, using multiple centres is crucial for the performance of \cite{DBLP:journals/corr/abs-1909-05235}. Fig. 4 of the work \cite{DBLP:journals/corr/abs-1909-05235} shows that when they use only $1$ centre per class, the performance drops by $3pp$, effectively making \cite{DBLP:journals/corr/abs-1909-05235} perform worse than the Group Loss by $2pp$. 

We further used the official code implementation to train their network using only one center on the  \textit{CARS 196} \cite{KrauseStarkDengFei-Fei_3DRR2013} dataset, reaching $83.1$ score in Recall@1, and $70.1$ score in NMI metric, with each score being $0.6pp$ lower than the score of The Group Loss. 
Essentially, when using the same backbone, SoftTriple loss \cite{DBLP:journals/corr/abs-1909-05235} reaches lower results than our method. 

As we have shown in the previous section, increasing the number of parameters improves the performances of the network, but it is not a property of the loss function. In fact, a similarly sized network to theirs (Densenet 169) consistently outperforms SoftTriple loss, as can be seen in Tab. \ref{tab:densenets}. 
For this reason, we keep this comparison in the supplementary material, while we leave for the main paper the comparisons with more than $20$ methods that use the same backbone.


\section{Alternative Loss Formulation}

In the main paper, we formulated the loss as an iterative dynamical system, followed by the cross-entropy loss function. In this way, we encourage the network to predict the same label for samples coming from the same class. 
One might argue that this is not necessarily the best loss for metric learning, in the end, we are interested in bringing similar samples closer together in the embedding space, without the need of having them classified correctly.
Even though several works have shown that a classification loss can be used for metric learning \cite{DBLP:conf/iccv/Movshovitz-Attias17,DBLP:journals/corr/abs-1811-12649,DBLP:journals/corr/abs-1909-05235}, we test whether this is also the best formulation for our loss function.

We therefore experiment with a different loss function which encourages the network to produce similar label distributions (soft labels) for the samples coming from the same class.
We first define Kullback-Leibler divergence for two distributions $P$ and $Q$ as:

\begin{equation}
    D_{KL}(P||Q) = \sum_{x \in X} P(x) log \frac{P(x)}{Q(x)}.
\end{equation}

We then minimize the divergence between the predicted probability (after the iterative procedure) of samples coming from the same class. Unfortunately, this loss formulation results in lower performances on both \textit{CUB-200-2011} \cite{WahCUB_200_2011} ($3pp$) and \textit{Cars 196} \cite{KrauseStarkDengFei-Fei_3DRR2013} ($1.5pp$). Thus, we report the experiments in the main paper only with the original loss formulation.

\section{Dealing with Negative Similarities}

Equation (4) in the main paper assumes that the matrix of similarity is non-negative.
However, for similarity computation, we use a correlation metric (see Equation (1) in the main paper) which produces values in the range $[-1, 1]$.
In similar situations, different authors propose different methods to deal with the negative outputs.
The most common approach is to shift the matrix of similarity towards the positive regime by subtracting the biggest negative value from every entry in the matrix \cite{DBLP:journals/neco/ErdemP12}.
Nonetheless, this shift has a side effect: If a sample of class $k_1$ has very low similarities to the elements of a large group of samples of class $k_2$, these similarity values (which after being shifted are all positive) will be summed up. If the cardinality of class $k_2$ is very large, then summing up all these small values lead to a large value, and consequently affect the solution of the algorithm.
What we want instead, is to ignore these negative similarities, hence we propose {\it clamping}. More concretely, we use a \textit{ReLU} activation function over the output of Equation (1).

We compare the results of shifting vs clamping. On the \textit{CARS 196} dataset, we do not see a significant difference between the two approaches.
However, on the \textit{CUBS-200-2011} dataset, the Recall@1 metric is $51$ with shifting, much below the $64.3$ obtained when using clamping.
We investigate the matrix of similarities for the two datasets, and we see that the number of entries with negative values for the \textit{CUBS-200-2011} dataset is higher than for the \textit{CARS 196} dataset. This explains the difference in behavior, and also verifies our hypothesis that clamping is a better strategy to use within {\it Group Loss}.

\section{Dealing with Relative Labels}
The proposed method assumes that the dataset consists of classes with (absolute) labels and  set of samples in each class. This is the case for the datasets used in metric learning \cite{WahCUB_200_2011,KrauseStarkDengFei-Fei_3DRR2013,DBLP:conf/cvpr/SongXJS16}) technique evaluations. However, deep metric learning can be applied to more general problems where the absolute class label is not available but only relative label is available. For example, the data might be given as pairs that are similar or dissimilar. Similarly, the data may be given as triplets consisting of anchor (A), positive (P) and negative (N) images, such that A is semantically closer to P than N. For example, in \cite{DBLP:conf/iccv/WangG15} a triplet network has been used to learn good visual representation where only relative labels are used as self-supervision (two tracked patches from the same video form a "similar" pair and the patch in the first frame and a patch sampled from another random video forms a "dissimilar" pair). Similarly, in \cite{DBLP:conf/eccv/MisraZH16}, relative labels are used as self-supervision for learning good spatio-temporal representation. 

Our method, similar to other classification-based losses \cite{DBLP:journals/corr/abs-1811-12649,DBLP:conf/iccv/Movshovitz-Attias17,DBLP:journals/corr/abs-1909-05235} for deep metric learning, or triple loss improvements like Hierarchical Triplet Loss \cite{DBLP:conf/eccv/GeHDS18} assumes the presence of absolute labels. Unlike traditional losses \cite{bromley1994signature,DBLP:conf/cvpr/SchroffKP15}, in cases where only relative labels are present, all the mentioned methods do not work. However, in the presence of both relative and absolute labels, then in our method, we could use relative labels to initialize the matrix of similarities, potentially further improving the performance of networks trained with The Group Loss.

\section{Dealing with a Large Number of Classes}

In all classification based methods, the number of outputs in the last layer linearly increases with the number of classes in the dataset. This can become a problem for learning on a dataset with large number of classes (say $N>1000000$) where metric learning methods like pairwise/triplet losses/methods can still work. In our method, the similarity matrix is square on the number of samples per mini-batch, so its dimensions are the same regardless if there are $5$ or $5$ million classes. However, the matrix of probabilities is linear in the number of classes. If the number of classes grows, computing the softmax probabilities and the iterative process becomes indeed computationally expensive. An alternative (which we have tried, reaching similar results to those presented in the paper) is to sparsify the matrix. Considering that in any mini-batch, we use only a small number of classes ($<10$), all the entries not belonging to these classes may be safely set to 0 (followed by a normalization of the probability matrix). This would allow both saving storage (e.g. using sparse tensors in PyTorch) and an efficient tensor-tensor multiplication. It also needs to be said, that in retrieval, the number of classes is typically not very large, and many other methods \cite{DBLP:journals/corr/abs-1811-12649,DBLP:conf/iccv/Movshovitz-Attias17,DBLP:journals/corr/abs-1909-05235,DBLP:conf/eccv/GeHDS18} face the same problem. However, in related fields (for example, face recognition), there could be millions of classes (identities), in which case we can use the proposed solution.

\section{t-SNE on CUB-200-2011 Dataset}

\begin{figure*}[h]
\centering
\includegraphics[width=0.96\textwidth, trim={1cm 3cm 0.5cm 5cm},clip]{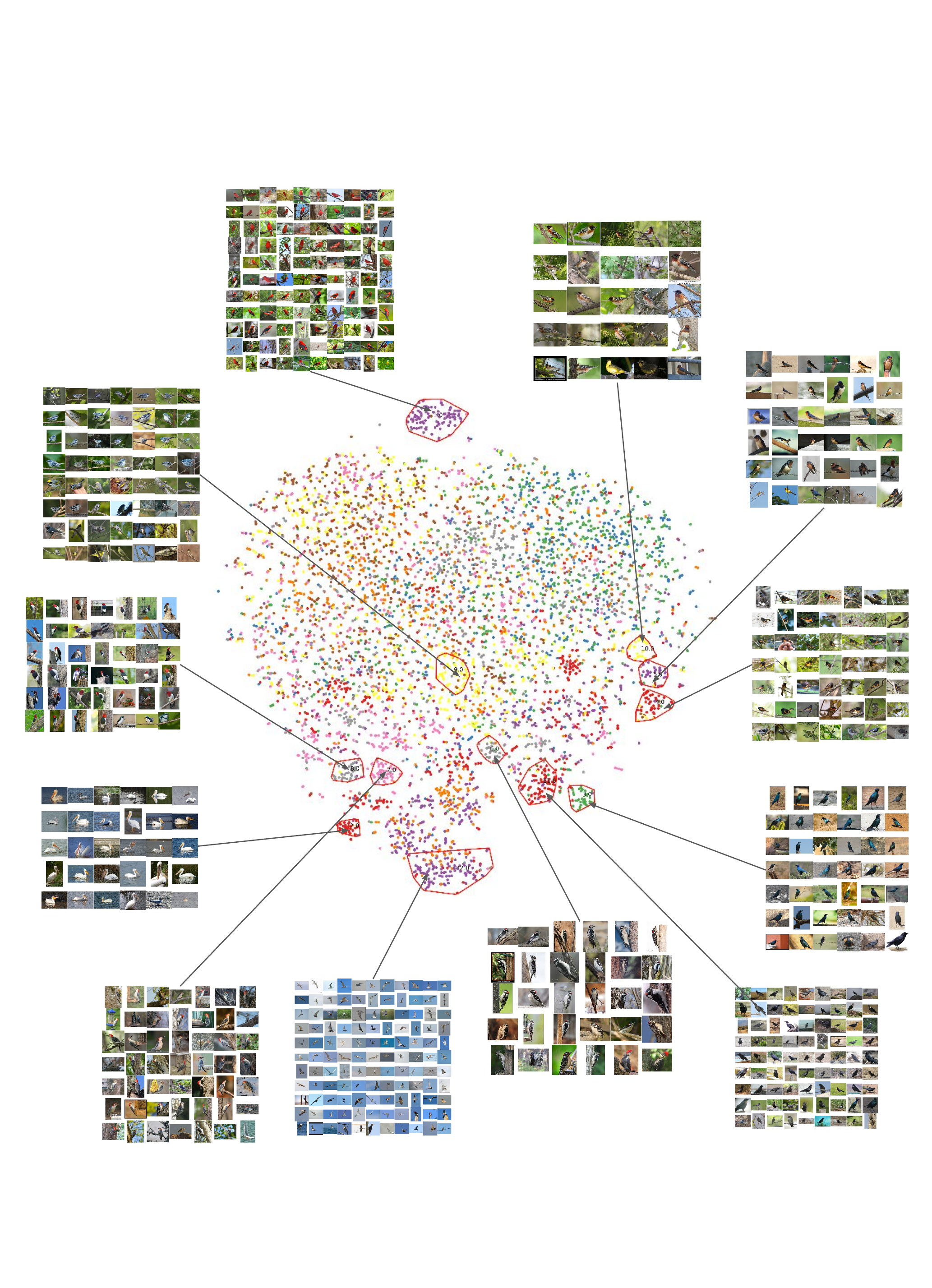}
  \caption{t-SNE \cite{DBLP:journals/ml/MaatenH12} visualization of our embedding
on the CUB-200-2011 \cite{WahCUB_200_2011} dataset, with some clusters highlighted. Best viewed on a monitor
when zoomed in.}
  \label{fig:t-sne}
\end{figure*}

Fig. \ref{fig:t-sne} visualizes the t-distributed stochastic neighbor embedding (t-SNE) \cite{DBLP:journals/ml/MaatenH12} of the embedding vectors obtained by our method on the \textit{CUB-200-2011} \cite{WahCUB_200_2011} dataset. 
The plot is best viewed on a high-resolution monitor when zoomed in. We highlight several representative groups by enlarging the corresponding regions in the corners. Despite the large pose and appearance variation, our method efficiently generates a compact feature mapping that preserves semantic similarity.

\clearpage
\newpage

%
%


%
%
\bibliographystyle{splncs04}
\bibliography{egbib}
\end{document}